\documentclass[12pt]{article}

\usepackage[margin=1in]{geometry}

\usepackage{authblk}

\usepackage[T1]{fontenc}
\usepackage{url}
\usepackage{hyperref}
\usepackage{enumitem}
\usepackage{sidecap}
\usepackage{comment}
\usepackage[numbers]{natbib}
\usepackage{adjustbox} 
\usepackage{natbib}

\usepackage{wrapfig}
\usepackage{graphicx}
\usepackage{dcolumn}
\usepackage{amsmath}
\usepackage{comment} 
\usepackage{amssymb}
\usepackage{mathrsfs} 
\usepackage[normalem]{ulem}
\usepackage{bm}
\usepackage{caption}
\usepackage{subfigure}
\usepackage{cleveref}

\usepackage[dvipsnames]{xcolor}

\usepackage{mathtools}
\renewcommand{\doteq}{\vcentcolon=}

\usepackage{cancel}
\usepackage{rotating}
\usepackage{listings}
\usepackage{tagging}
\usepackage{hyphenat}
\usepackage{algorithm}
\usepackage{algpseudocode}
\usepackage{multirow}
\usepackage{amsthm}
\theoremstyle{definition}

\theoremstyle{definition}

\newtheorem*{exercise*}{Exercise}
\theoremstyle{remark}

\newtheorem*{remark}{Remark}

\graphicspath{{./figures/}}

\title{U-Turn Diffusion}

\author{Hamidreza Behjoo \& Michael Chertkov}

\affil{University of Arizona, Tucson, AZ 85721, USA \authorcr Email: \texttt{[hbehjoo,chertkov]@arizona.edu} 
}

\begin{document}

\maketitle

\begin{abstract}

We investigate diffusion models generating synthetic samples from the probability distribution represented by the Ground Truth (GT) samples. We focus on how GT sample information is encoded in the Score Function (SF), computed (not simulated) from the Wiener-Ito (WI) linear forward process in the artifical time $t\in [0\to \infty]$, and then used as a nonlinear drift in the simulated WI reverse process with $t\in [\infty\to 0]$. We propose \textbf{U-Turn diffusion}, an augmentation of a pre-trained diffusion model, which shortens the forward and reverse processes to $t\in [0\to T_u]$ and $t\in [T_u\to 0]$. The U-Turn reverse process is initialized at $T_u$ with a sample from the probability distribution of the forward process (initialized at $t=0$ with a GT sample) ensuring a detailed balance relation between the shorten forward and reverse processes. Our experiments on the class-conditioned SF of the ImageNet dataset and the multi-class, single SF of the CIFAR-10 dataset reveal a critical \textbf{Memorization Time} \( T_m \), beyond which generated samples diverge from the GT sample used to initialize the U-Turn scheme, and a \textbf{Speciation Time} \( T_s \), where for \( T_u > T_s > T_m \), samples begin representing different classes. We further examine the role of SF non-linearity through a Gaussian Test, comparing empirical and Gaussian-approximated U-Turn auto-correlation functions, and showing that the SF becomes effectively affine for \( t > T_s \), and approximately affine for $t\in [T_m,T_s]$.

\end{abstract}

\section{Introduction}

The fundamental mechanics of Artificial Intelligence (AI) involve a three-step process: acquiring Ground Truth (GT) data, modeling it, and then predicting or inferring based on the model. The core component of the prediction phase of the Generative AI (GenAI) is the generation of Synthetic Data (SD).

The success of a GenAI model depends on how effectively it embeds information about SD. In the emerging landscape of GenAI, this is achieved by leveraging the rich structure within these models. In particular, Score-Based Diffusion (SBD) models \cite{song_score-based_2021, Rombach_2022_CVPR, ho_denoising_2020} have emerged as a highly successful paradigm, currently representing the state of the art in image generation and competitive in other applications. SBD models excel due to their inherent structure, which consists of forward and reverse stochastic dynamic processes in an auxiliary time, enabling them to extract and redistribute information over time.

Remarkably, the principles underlying SBD models date back to earlier ideas from Stochastic Differential Equations (SDEs) by Anderson \cite{anderson_reverse-time_1982}. These principles allow for the design of a reverse process where the marginal probability distributions of images in the forward (noise-adding) and reverse (de-noising) processes are identical by construction, and moreover preserve a Detailed Balance (DB) relation between the forward and reverse processes. This theoretical construct serves as a guiding principle for SBD models, enabling the redistribution of information from GT data, which forms the initial condition for the forward SDE. Over time, this information is encoded into the Score Function (SF) representing the re-scaled drift term in the reverse SDE, facilitating the generation of high-quality SD. Neural Networks (NNs) are employed to fit the exact SF, derived from Anderson's work, for efficiency, avoiding memorization, and providing smoother representations dependent on the GT data.

\subsection{Our Contributions}

The concept of the U-Turn diffusion augmentation of a pre-trained Score-Based Diffusion (SBD) model arises from the intuition that the success of the SBD methodology is due to the reverse Stochastic Differential Equation (SDE) containing all necessary information to generate synthetic data. It also stems from addressing key questions essential to refining this approach: How is information about the GT samples distributed over time within the Score Function (SF), which represents the drift term in the reverse SDE? Can we rigorously shorten the theoretically infinite time interval while preserving Anderson's relations between statistics of the forward and reverse processes? How does the nonlinearity of the score function evolve over time? In the remainder of this subsection, we detail how this manuscript addresses each of these questions.

We investigate diffusion models of generative AI, focusing on generating new synthetic images from a probability distribution representing Ground Truth (GT) samples. A typical diffusion model consists of a forward noise-injecting phase, modeled as a linear Ito-Wiener stochastic process, which allows for analytical computation, and a reverse denoising phase, also an Ito-Wiener process but with a nonlinear drift term, requiring simulation to produce synthetic images. Our work centers on understanding how information about the GT samples, used to initiate the forward process, is encoded in the SF, which is approximated by a NN. The SF is crucial because it links the reverse process to the forward process: it is defined as the gradient of the log of the marginal probability distribution in the forward process at each point in time and is subsequently used as the drift term in the reverse process.

The main observation and contribution of this manuscript is that the essential information about the GT samples is primarily encoded in the SF during the initial stages of the forward process. This insight leads us to propose the {\bf U-Turn diffusion} model \footnote{The original version of this manuscript was reported on arXiv in August 2023.}, which revises the original diffusion model by shortening both the forward process and the subsequent reverse dynamics. In the U-Turn model, the reverse process becomes a detailed-balance conjugate of the forward process, requiring it to start from a sample of the marginal probability distribution at the forward process's final, `U-Turn' time, where the forward process is initialized with a GT sample. Notably, this construction allows for the creation of the U-Turn model directly from the original diffusion model's score function without the need for retraining. While the U-Turn algorithm itself is described in Section \ref{sec:U-Turn}, it is also prepared via a preliminary discussion, which can be viewed as a technical introduction into diffusion models, presented in Section \ref{sec:stage}. 

Our second contribution, reported in Section \ref{sec:basic-SBD}, is development of a toolbox of tests -- the Kolmogorov- Smirnov Gaussianity test, score-function amplitude test, and U-Turn auto-correlation tests -- to determine appropriate U-Turn time. These tests complement the standard Fréchet Inception Distance (FID) tests used to evaluate quality of the difussion models. 

\begin{figure}[H]    
\subfigure[ImageNet]{ 
        \centering
        \includegraphics[scale=0.53]{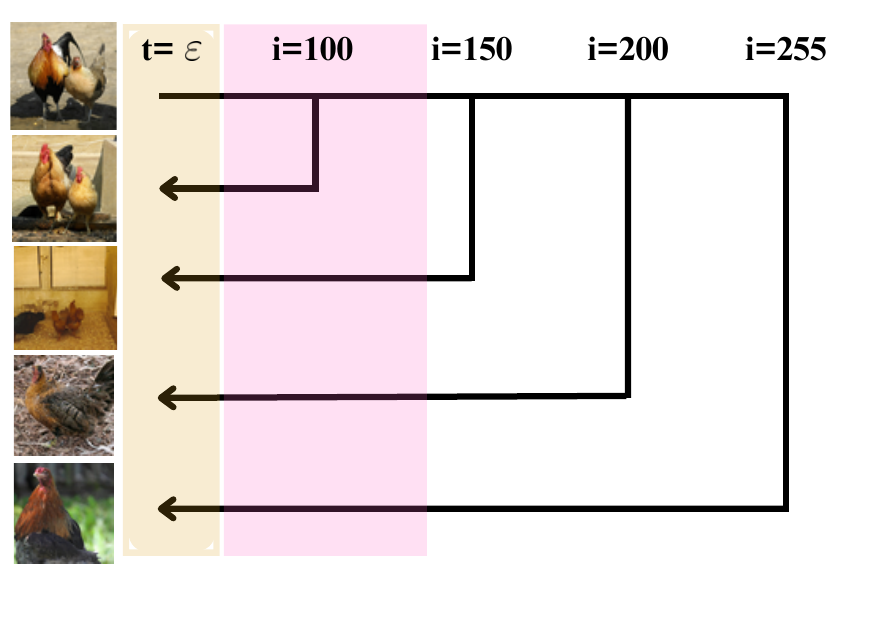} 
        }
    \subfigure[CIFAR-10]{
        \centering
        \includegraphics[scale=0.48]{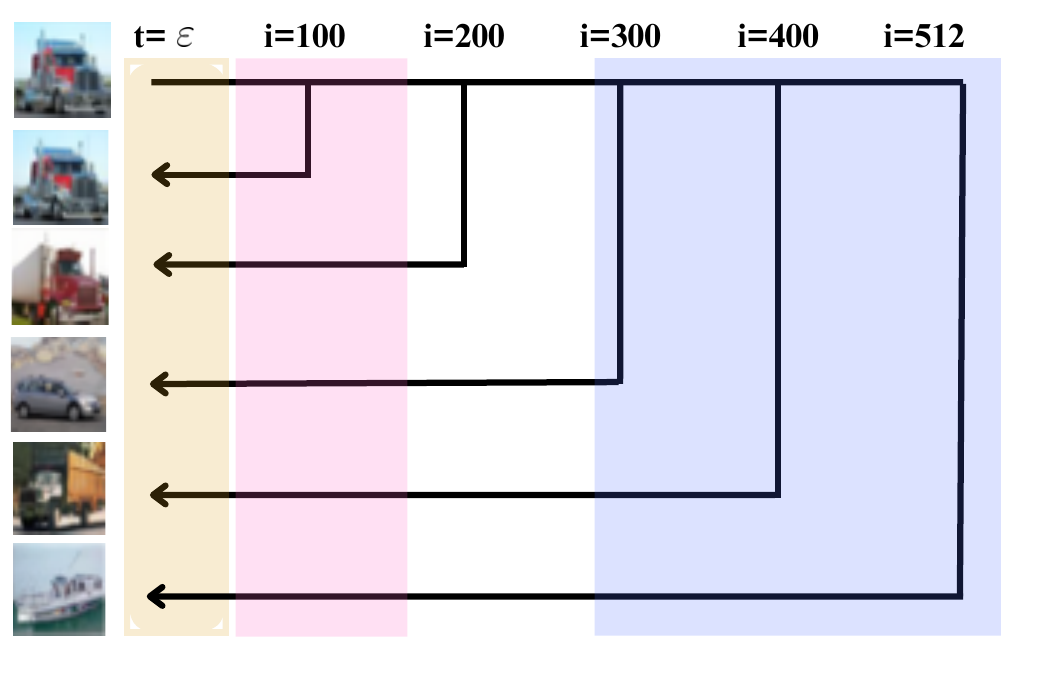} }
    \caption{Illustration of the U-Turn concept for (a) ImageNet and (b) CIFAR-10. Our analysis, based on newly introduced tests, reveals that making the U-Turn earlier is beneficial, but not too early. The dark-yellow region indicates a small vicinity near the origin where the approximation of the score function by a NN is crucial to avoid memorization (i.e., generation of ground truth samples). The pink regions mark the range where memorization transitions, \(T_m\), occur. These transitions are observed in both single-class (ImageNet) and multi-class settings. In the multi-class case (with a single score function for the entire dataset), we also observe the speciation transition, \(T_s\), previously reported in \cite{biroli_dynamical_2024} and depicted in light blue in the right figure. This schematic illustration emphasizes an important observation of this work: both \(T_m\) and \(T_s\) are not self-averaged. Instead, they fluctuate across GT samples and even between different realizations of the forward process -- thus resulting in ranges (distributions).}
\label{fig:scales}
\end{figure}

Third, and as reported in Section \ref{sec:U-Turn}, we test the U-Turn scheme, as well as the aforementioned toolbox of tests on ImagenNet datasets. Our experiments with the pre-computed/trained Score Function (SF) of the class-conditioned ImageNet dataset \cite{karras_elucidating_2022} lead to the discovery of a critical U-Turn time, \( T_m \), which we refer to as the \textbf{Memorization Time}. Specifically, if the U-Turn time, \( T_u \), is shorter than \( T_m \), the new sample generated is close to the original GT sample used to initialize the reverse process at \( T_u \); however, if \( T_u > T_m \), a truly new sample is produced. We observe that \( T_m \) varies across different classes within the ImageNet dataset. The schematic illustration of what U-Turn discovers in the case of a single-class score function in illustrated in Fig.~(\ref{fig:scales} a). 

Fourth, we conduct in Section \ref{sec:CIFAR} experiments with the multi-class CIFAR-10 dataset and its pre-trained averaged over the entire dataset (that is not class specific) Score Function (SF) from \cite{karras_elucidating_2022}. Here, in addition to observing the {\bf Memorization Transition}, we identify a \textbf{Speciation Transition} \cite{biroli_dynamical_2024} at time \( T_s \), where \( T_s > T_m \). When the U-Turn time \( T_u \) exceeds \( T_s \), the resulting image is not only distinct from the initial image but also belongs to a different class than the initial image. The schematic illustration of the U-Turn shows in the case of a multi-class score function in illustrated in Fig.~(\ref{fig:scales} b). 

The next two contributions, in particular the fifth one also discussed in Section \ref{sec:CIFAR}, came from the follow up experiments with the averaged-over-the-multi-class score-function of the CIFAR-10 dataset. By inspecting synthetic samples we observe that both $T_m$ and $T_s$ fluctuate rather significantly from class to class and even from sample-to-sample. This motivated us to analyze the U-Turn auto-correlation function, however now conditioned to a single GT sample and to a single class. We observe, now quantitatively, that the GT-sample constrained U-Turn auto-correlation function fluctuates significantly from sample to sample,  that is (using statistical physics terminology) not a self-averaged quantity. This lack of self-averagness persists, even though it became weaker, when we compare U-Turn auto-correlation functions (conditioned to a class of the GT sample) for different classes.

Six, we pose in Section \ref{sec:Gauss} the question whether the dynamic phase transitions at \( T_m \) and \( T_s \) are related to the non-linearity of the Score Function (SF). To investigate this, we develop a Gaussian Test, and related G-Turn algorithm, assuming the GT data is Gaussian or can be approximated by a Gaussian distribution, which then allows for an analytical expression of the U-Turn auto-correlation function. We observe that, while the empirical U-Turn auto-correlation functions evaluated on the ImageNet dataset and  on  the averaged-over-all-the-classes CIFAR-10 dataset are relatively close to the corresponding estimate for the U-Turn auto-correlation function computed within the Gaussian ansatz, there is still a notable divergence between them. A detailed examination of the U-Turn sampling procedure in the CIFAR-10 experiments leads to a key final observation: the score function becomes effectively linear for \( T_u > T_s \), and that it is approximately linear, however still with notable deviations from linearity detected via the FID test.

We also discuss empirical results of applying the U-Turn diffusion to the case of deterministic sampler (reversed process) in Section \ref{sec:deterministic-samplers}.

We conclude the paper in Section \ref{sec:conclusions} by summarizing the results and sharing some path forward ideas.

\subsection{Related Work}\label{sec:related-work}

In \cite{raya2023spontaneous}, the authors examined diffusion models from the perspective of symmetry breaking, revealing that the early stages of the reverse dynamics contributed minimally to generating new samples. Based on these observations, they proposed a "late start" strategy for the reverse process, which is similar to the U-Turn discussed in this paper. We consider this work and two other papers discussed next as foundational to our more systematic analysis presented here.

In \cite{biroli_dynamical_2024} the authors -- motivated by the analysis of a toy example where the GT data is sampled from a bimodal distribution represented by a mixture of two Gaussians reported in \cite{biroli_generative_2023} -- reported emergence of the dynamic phase transitions of the memorization and speciation type. Even though memorization transition was reported earlier in the first version of our manuscript, our speciation transition analysis on the multi-class model as well as analytic analysis of the diffusion in the case of the Gaussian GT data reported in this manuscript were  inspired by respective analysis of \cite{biroli_generative_2023,biroli_dynamical_2024}. 

Phase transition methodology of statistical physics also became central to the "forward-reversed experiments" of \cite{sclocchi2024phase}, where the authors experimented with a setting similar to the U-Turn. They empirically reported a phase transition in the value of $T_u$ and provided some statistical physics estimations based on a mean-field approximation.

Another related method, called "boomerang," was reported in \cite{luzi_boomerang_2022}. This method focuses on generating images that closely resemble the original samples, whereas U-Turn diffusion aims to create distinctly different images that approximate i.i.d. samples from the entire dataset.

\section{Technical Introduction: Score-Based Diffusion}
\label{sec:stage}

We follow the Score-Based Diffusion (SBD) framework, as expounded in \cite{song_score-based_2021}. The SBD harmoniously integrates the principles underlying the "Denoising Diffusion Probabilistic Modeling" framework introduced in \cite{sohl-dickstein_deep_2015} and subsequently refined in \cite{ho_denoising_2020}, along with the "Score Matching with Langevin Dynamics" approach introduced by \cite{NEURIPS2019_song}. This seamless integration facilitates the reformulation of the problem using the language of stochastic differential equations, paving the way to harness the  Anderson's Theorem \cite{anderson_reverse-time_1982}. As elucidated in the following, this theorem assumes a principal role in constructing a conduit linking the forward and reverse diffusion processes.

Let us follow \cite{song_score-based_2021} and introduce the forward-in-time SDE for the vector ${\bm x}_t=(x_{t;i}|i=1,\cdots,d)\in \mathbb{R}^d$ (where $d$ is the embedding space for the data):
\begin{align}
    \label{eq:forward_sde}
    & \text{\underline{Forward}:}\ t\in [0\to T]:\quad d{\bm x}_t = {\bm f}({\bm x}_t, t)dt+{\bm g}({\bm x}_t, t)d{\bm w}_t,
\end{align}
and another reverse-in-time SDE:
\begin{align}\nonumber 
    \text{\underline{Reversed}:}\ t\in [T\to 0]:\ d{\bm y}_t  & =\left({\bm f}({\bm y}_t, t )-\! {\nabla}\cdot {\bm G}({\bm y}_t, t)- {\bm G}({\bm y}_t,t ) s( {\bm y}_t, t) \right)dt\\ & + \ {\bm g}({\bm y}_t,t)d\Bar{{\bm w}}_t, \label{eq:reverse_sde}
\end{align}
where the drift/advection ${\bm f} : \mathbb{R}^{d} \times \mathbb{R} \rightarrow{} \mathbb{R}^{d}$ and diffusion ${\bm g} : \mathbb{R}^{d} \times \mathbb{R}\rightarrow \mathbb{R}^{d} \times \mathbb{R}^{d}$ are sufficiently smooth (Lipschitz functions); and ${\bm G}({\bm y}_t, t) = {\bm g}({\bm y}_t, t) {\bm g}({\bm y}_t, t)^\top$. Here, $s({\bm x}_t,t)\doteq \nabla_{{\bm x}_t} \log p({\bm x}_t,t)$ is the so-called score-function  computed for the marginal probability distribution of the forward process (\ref{eq:forward_sde}) and utilized in Eq.~(\ref{eq:reverse_sde}) to drive the reverse process. 
Both forward and reversed processes are subject to Ito-regularization. ${\bm w}_t$ and $\Bar{{\bm w}}_t$ represent standard Wiener processes for forward and reverse in time, respectively.

The forward diffusion process transforms the \textit{initial distribution} $p_{\text{data}}(\cdot)$, represented by samples, into a \textit{final distribution} $p_T(\cdot)$ at time $T$. The terms ${\bm f}({\bm x}_t,t)$ and ${\bm g}({\bm x}_t,t)$ in the forward SODE (\ref{eq:forward_sde}) are free to choose, but in the SBD approach, they are usually selected in a data-independent manner. 

Anderson's theorem establishes that the forward-in-time process  and the reverse-in-time process have the same marginal probability distribution, provided that the reverse process is initiated at $t=T$ with the marginal probability distribution identical to the marginal forward probability distribution evaluated at $t=T$ \footnote{Note that Anderson actually proved a stronger result in \cite{anderson_reverse-time_1982}, establishing equivalence not only between the marginal probabilities but also transition probabilities of the general forward process (\ref{eq:forward_sde}) and the reversed process (\ref{eq:reverse_sde}). This relationship is akin to the Detailed Balance (DB) condition in equilibrium statistical mechanics, however extended here to non-autonomous processes. Additionally, there are approaches, notably the so-called probability flow method of \cite{song_score-based_2021}, where the reversed process is deterministic, and the DB condition for the transition probabilities is broken, however, the marginal probability distributions of the forward and reversed processes are still equal to each other. The authors are grateful to G. Birolli and M. Mézard for their valuable discussions that helped to clarify this point.}.
\begin{remark}
    The proof of Anderson's Theorem relies on the equivalence of the Fokker-Planck equations derived for the direct (\ref{eq:forward_sde}) and inverse (\ref{eq:reverse_sde}) dynamics:
    \begin{align} \label{eq:FP-direct}
        & \partial_t p+\nabla_i\left(f_i p\right) & =\ \ \frac{1}{2}\nabla_i\nabla_j\left(({\bm g} {\bm g}^T)_{ij}p\right),\\
        \label{eq:FP-inverse}
        & \partial_t p+\nabla_i\left(f_i p\right) -\nabla_i\left(p \nabla_j ({\bm g}{\bm g}^T)_{ij}\right)-\nabla_i \left({\bm g}{\bm g}^T)_{ij} s_j p\right) &=-\frac{1}{2}\nabla_i\nabla_j\left(({\bm g}{\bm g}^T)_{ij}p\right),
    \end{align}
    where $i,j=1,\cdots,n_x$ and we assume summation over repeated indexes; and we also dropped  dependencies on ${\bm x}$ or ${\bm y}$ and on $t$ in ${\bm f}, {\bm g}$ and ${\bm s}$ to make the notations lighter.
\end{remark}

{\bf Inference}, which involves generating new samples from the distribution represented by the data, entails initializing the reverse process (\ref{eq:reverse_sde}) at a sufficiently large (but practically finite) $t=T$ with a sample drawn from the normal distribution derived in the limit of $T\to\infty$, $p$
and then running the process reversed in time to reach the desired result at $t=0$. This operation requires accessing the
Score Function (SF), 
as indicated in Eq.~(\ref{eq:reverse_sde}). However, practically obtaining the exact time-dependent SF is challenging. Therefore, we normally resort to approximating it with a Neural Network (NN) parameterized by a vector of parameters $\theta$: ${\bm s}_{\theta}(\cdot,t) \approx {\bm s}(\cdot,t)$.

The NN-based approximation of the SF allows us to efficiently compute and utilize gradients with respect to the input data ${\bm x}$ at different times $t$, which is essential for guiding the reverse process during inference. By leveraging this NN approximation, we can effectively sample from the desired distribution and generate new images which are approximately i.i.d. from a target probability distribution represented by input data. This approach enables us to achieve reliable and accurate inference in complex high-dimensional spaces, where traditional methods may struggle to capture the underlying data distribution effectively.

{\bf Training:} The NN ${\bm s}_\theta(\cdot,t)$ can be trained to approximate\\ ${\bm s}({\bm x}_t)=\nabla_{{\bm x}_t}{\log {p_t({\bm x}_t)}}$ using, for example, the weighted De-noising Score Matching (DSM) objective \cite{song_score-based_2021}:
\begin{gather*} 
    \mathbb{E}_{  t \sim U(0,T), {\bm x}_0 \sim p_0(\cdot), {\bm x}_t \sim p_t(\cdot | {\bm x}_0) }   \left[\frac{\lambda(t)}{2} \|\nabla_{{\bm x}_t}{\log {p_t({\bm x}_t | {\bm x}_0)}} - {\bm s}_\theta({\bm x}_t,t)\|_2^2\right].  
\end{gather*} 
In the experiments presented in this manuscript, we do not train the models ourselves. Instead, we leverage pre-trained models (score functions) from existing open-source work specifically \cite{karras_elucidating_2022}, for class conditioned ImageNet, which provides a separate score function for each class, and for multi-class CIFAR-10, which uses a single score function for the entire dataset. Our focus is primarily on analyzing and understanding the basic Score-Based Diffusion (SBD) scheme, and subsequently on improving it by proposing the U-Turn scheme and related modifications.

\subsection{Choice of SBD  -- Time-Dependent Brownian Diffusion}

In this manuscript we choose to work with the simplest SBD model --- the time-dependent Brownian diffusion: the drift term ${\bm f}({\bm x}_t, t)$ is zero and the diffusion term ${\bm g}({\bm x}_t, t)$ in space-independent but time-dependent -- $\sqrt{2\beta_t}$. This results, according to Eq.~(\ref{eq:forward_sde}), in the following explicit expression for the exact SF:
\begin{gather}\label{eq:s-exact}
{\bm s}({\bm x}_t,t)=\nabla_{\bm x}\log\left(\sum\limits_{n=1}^N {\cal N}\left({\bm x}_t|{\bm x}^{(n)};2\hat{\bm I}\int_0^t dt' \beta_{t'}\right)\right),
\end{gather}
where $n=1,\cdots,N$ indexes the GT samples, $\hat{\bm I}$ is the identity matrix; and ${\cal N}\left({\bm x}|{\bm \mu};\hat{\bm \Sigma}\right)$ is the normal distribution of ${\bm x}$ with mean vector ${\bm \mu}$ and covariance matrix $\hat{\bm \Sigma}$.

Two key remarks are warranted regarding the universality of the SBD approach and the selection of the SBD model.

First, we highlight that explicit expressions for the score function (SF), such as the one presented in Eq.~(\ref{eq:s-exact}), are a common feature across various SBD models. This explicit form of the SF eliminates the need to simulate the forward process, offering a significant computational advantage.

Second, an important observation in the field is that, while the details of the forward model are critical for practical implementation, the choice of the underlying model is surprisingly flexible. For instance, the basic model in Eq.~(\ref{eq:s-exact}), which we employ in our experiments, reliably supports the generation of high-quality images across diverse settings.

We now turn to a discussion of the discretization and the selection of the $\beta$-protocol, addressing each in the context of the two pre-trained use cases developed in \cite{karras_elucidating_2022}.

\subsubsection*{Pre-Trained ImageNet-64 and CIFAR-10}

For the ImageNet-64 dataset \footnote{https://www.image-net.org/}, the total duration of the SBD process was set to \( T = 80 \), with the time-dependent diffusion coefficient defined as \(\beta(t) = 2t\). The interval \([0, T]\) was discretized into 255 non-uniform time steps, denoted as \(\{t_i\}\). Larger time steps were allocated during the early stages of the reversed SDE and progressively smaller steps in the later stages. Further details on this discretization can be found in Appendix D.1 of \cite{karras_elucidating_2022}. In this setup, the time dependence \( t = t_i \) is indexed as \( i = 0, \cdots, 255 \).

We utilized the open-source data and code from \cite{karras_elucidating_2022}, which eliminated the need to retrain the model to generate \({\bm s}_{\bm \theta}({\bm x}_t; t)\), a neural network (NN) approximation of Eq.~(\ref{eq:s-exact}). This significantly reduced computational overhead. The pre-trained model achieved a Fréchet Inception Distance (FID) score of 1.36 using 511 Neural Function Evaluations (NFE), corresponding to evaluating the NN version of Eq.~(\ref{eq:s-exact}) twice per discretization step. In our implementation, we observed a slightly higher FID score of 1.42 for 50,000 generated images. Although this represents a marginal difference, it highlights the robustness of the original method and the reproducibility of the results reported in \cite{karras_elucidating_2022}.

For the CIFAR-10 dataset \footnote{https://www.cs.toronto.edu/~kriz/cifar.html}, a similar SBD process was applied with \( T = 80 \) and \(\beta(t) = 2t\). The time interval \([0, T]\) was discretized into 512 non-uniform steps \(\{t_i\}, i = 1, \cdots, 512\), with larger steps in the early stages of the reversed SDE and smaller steps in the later stages, following the approach of \cite{karras_elucidating_2022}. The pre-trained model achieved an FID score of 3.5 with 1024 NFE, where each discretization step involved evaluating the NN version of Eq.~(\ref{eq:s-exact}) twice. In our implementation, the FID score was slightly higher at 3.65 for 50,000 generated images. This small deviation again underscores the robustness of the original approach and the reproducibility of the results.

\section{Analysis of Basic SBD Model} \label{sec:basic-SBD}

In the following, and as custom in the field, we will extensively use the Fréchet Inception Distance (FID) metric to evaluate the quality of images generated by basic diffusion models and their U-Turn counterparts. FID measures the similarity between the distributions of ground truth (GT) and generated images by approximating them as normal distributions. (See Appendix \ref{sec:FID} for details.) Although the FID score does not assess whether the generated images are correlated with specific GT images, this approach is justified in the context of standard SBD modeling. In such models, the reverse process is initialized independently and identically distributed (i.i.d.) from a normal distribution, preventing memorization of GT samples.

However, this logic does not apply to our task of analyzing how the ensemble of initial GT images, or a particular initial GT image, becomes forgotten over time as we advance with the forward SDE. To address this limitation in the analysis of SBD models, we introduce additional tests alongside the FID test: the Kolmogorov-Smirnov (KS), and norm of SF test. These tests are discussed and then applied in the following subsections to analyze the SBD models over 1,000 generated samples of ImageNet, which serves as our first and class-specific/fixed working example. Later in the manuscript we will also discuss application of the tests to the multi-class CIFAR-10 data.

\subsection{Kolmogorov-Smirnov Test} \label{sec:KS-main}

\begin{figure}[H]
    \centering
    \includegraphics[scale=0.65]{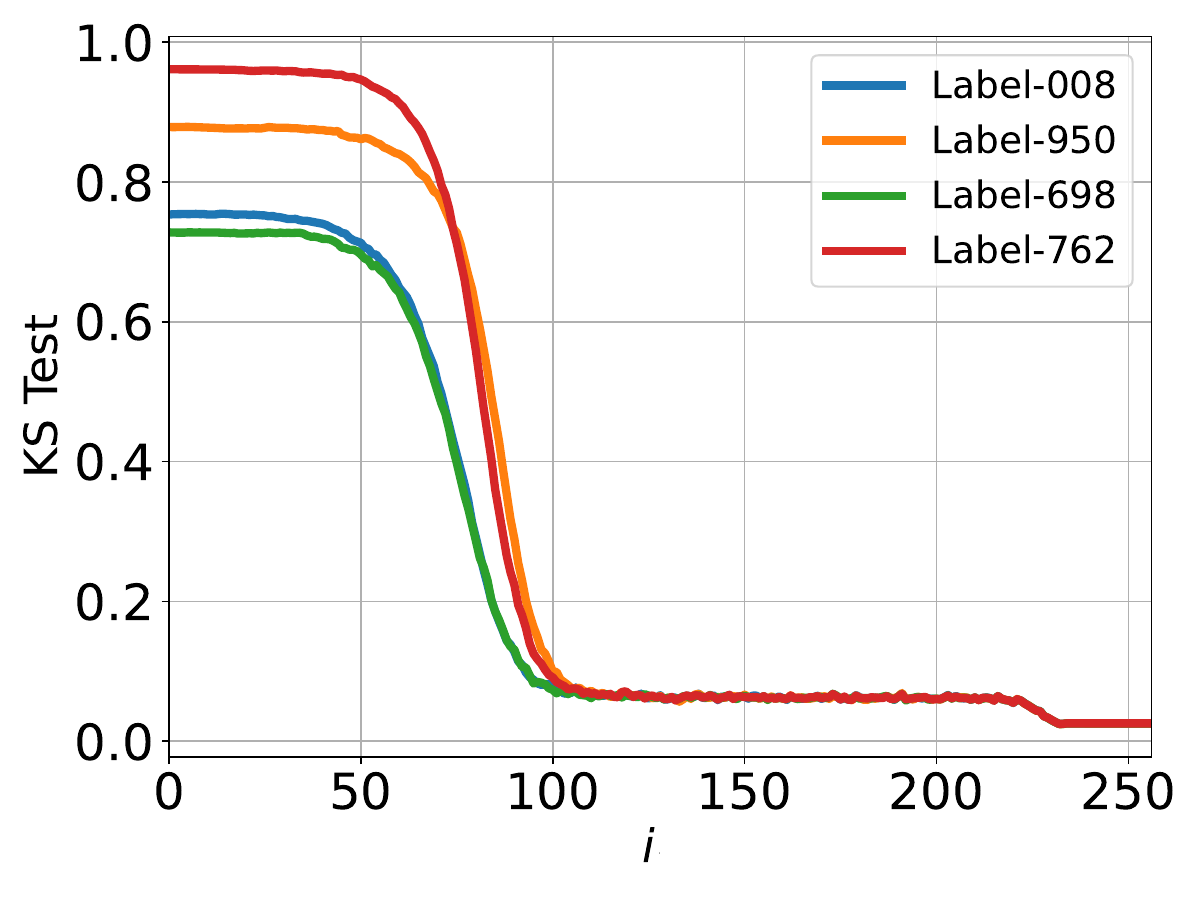}
    \caption{Kolmogorov-Smirnov (KS) test applied to the ImageNet dataset. Different colors represent different classes/labels: 008 (hen), 950 (orange), 698 (palace), and 762 (restaurant, eating house, eatery).}
    \label{fig:ks_test_many_class_ImageNet}
\end{figure}

We employ the Kolmogorov-Smirnov (KS) Gaussianity test to evaluate the null hypothesis: "Is a single-variable marginal of the resulting multivariate distribution of ${\bm x}(t)$ at a given time $t$ Gaussian?" To address this hypothesis, the KS test is applied to each single-variable marginal, defined as \( p_t(x_k) = \int d({{\bm x}} \setminus x_k) p_t({{\bm x}}) \). For implementation details, refer to Appendix \ref{sec:KS}.

The results of the KS analysis for the reverse process in the ImageNet experiment are shown in Fig.~\ref{fig:ks_test_many_class_ImageNet}. 

We observe a smooth yet significant transition in the KS factor over time, starting around \( i \approx 50 \), where \( p_t({\bm x}_t) \) is far from Gaussian, and progressing to \( i \approx 100 \), where the distribution becomes much closer to Gaussian. This transition suggests that the process of encoding information from the ground truth (GT) samples into the time- and GT-sample-dependent score function is largely completed by \( i \approx 100-200 \).

Interestingly, the initial level of non-Gaussianity differs across labels at \( i = 0 \). However, as time progresses, the KS curves for different labels converge, becoming nearly identical. This indicates a uniform Gaussianization process across labels as the reverse process evolves.

Additionally, further "Gaussianization" is observed at \( i \approx 225 \), as evidenced by a decrease in the KS factor. It is important to note that the KS test evaluates only the Gaussianity of spatial marginals—associated with a single component of \( {\bm x} \). This restricted Gaussianity does not necessarily imply that the entire vector \( {\bm x} \) follows a Gaussian distribution.

We also explored various bivariate KS tests, but these exhibited significant variability across tests. This variability motivates the exploration of additional quantitative methods, which are discussed in subsequent sections. The effective Gaussianity of the forward process at later times and reversed process at early time will be revisited in Section \ref{sec:Gauss} and Appendix \ref{app:Brownian}.

\subsection{Average of the Normalized Score Function 2-Norm} \label{sec:SFN}

\begin{figure}[H]
    \centering
    \includegraphics[scale=0.65]{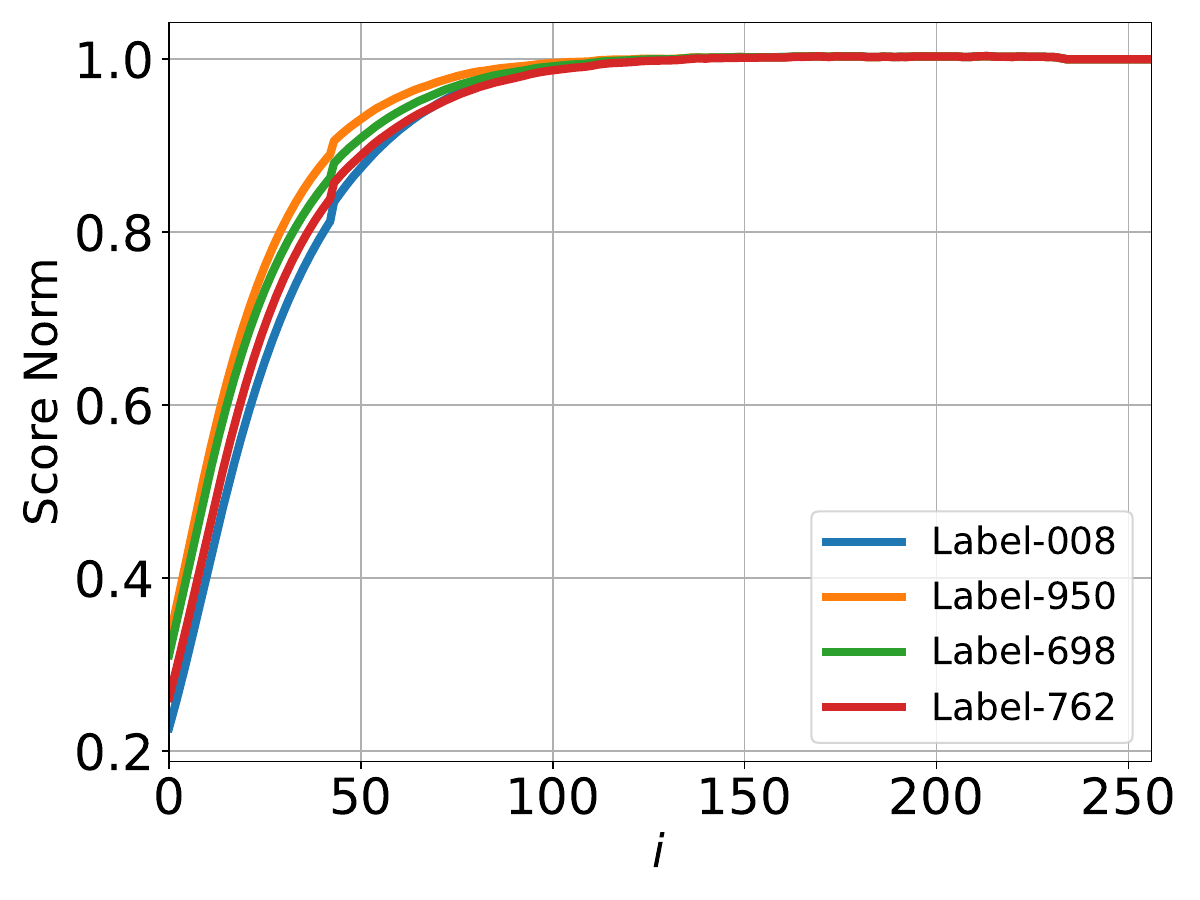}
    \caption{Average of the normalized score function 2-norm test for ImageNet. Consistent with Fig.~\ref{fig:ks_test_many_class_ImageNet}, different colors represent different classes/labels.}
    \label{fig:norm_test_ImageNet}
\end{figure}

The time dependence of the normalized 2-norm of the score function, averaged over multiple paths or instances of the reversed process, is presented in Fig.~\ref{fig:norm_test_ImageNet}. Details of this computation can be found in Appendix \ref{sec:score-norm}. 

Consistent with the dynamics observed in the Kolmogorov-Smirnov (KS) analysis (Fig.~\ref{fig:ks_test_many_class_ImageNet}), significant changes in the normalized score function 2-norm begin at approximately \( i \approx 50 \) and are largely completed by \( i \approx 100-150 \). This behavior suggests that, if the score function norm is used as a criterion, no substantial additional information about the ground truth (GT) samples is being encoded into the score function at later times. 

These observations align with the trends and discussions presented in subsequent sections, further supporting interpretation of different stages in the reverse process dynamics.

\subsection{Insensitivity to Reverse Process Initialization}

\begin{figure}[H]
    \centering
    \begin{tabular}{lc}
        \begin{turn}{90} \quad \tiny Gaussian \end{turn} &
        \includegraphics[scale=0.5]{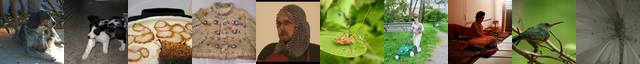} \\
        
        \begin{turn}{90} \quad \tiny Uniform \end{turn} &
        \includegraphics[scale=0.5]{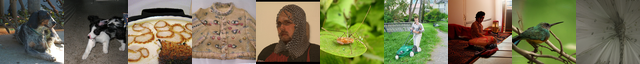} \\
        
        \begin{turn}{90} \quad \tiny Bernoulli \end{turn} &
        \includegraphics[scale=0.5]{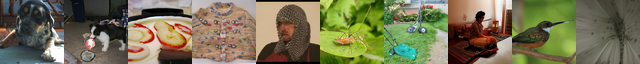} \\
        
        \begin{turn}{90} \quad \tiny Zero \end{turn} &
        \includegraphics[scale=0.5]{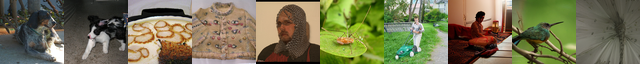} \\
        
        \begin{turn}{90} \quad \tiny Data \end{turn} &
        \includegraphics[scale=0.5]{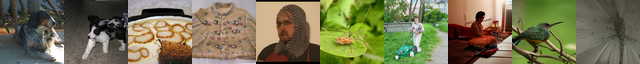} \\
    \end{tabular}
    \caption{Results of running the reverse process with different initializations: Gaussian (\(x_T \sim \mathcal{N}(0,I)\)), Uniform (\(x_T \sim \text{Uniform}[-1,1]\)), Zero (\(x_T = 0\)), GT Data (\(x_T \sim p_{\text{data}}\)), and Bernoulli (\(x_T \sim p_{\text{Bernoulli}}\)).}
    \label{fig:class_cond}
\end{figure}

In standard SBD modeling, the reverse process is typically initialized with an i.i.d. sample drawn from the probability distribution of the forward process at time \( T \), or at \( T \to \infty \) if such a limit exists. This distribution is often a simple Gaussian. However, as demonstrated in Fig.~\ref{fig:class_cond}, the specifics of the reverse process initialization appear to have minimal impact on the generated output samples. The figure shows results obtained from the same model but initialized with different distributions.

This insensitivity to the specifics of reverse process initialization, combined with the early temporal saturation observed in both the KS and score function norm tests discussed in previous subsections, suggests the potential to shorten the durations of both the forward and reverse processes. Motivated by this observation, we shift our focus to exploring this approach in the next section.

\section{U-Turn}\label{sec:U-Turn}

Motivated by the analysis in the preceding section, we introduce a modification of the basic SBD process, which we call U-Turn. Our analysis of the dynamics of the basic SBD process, particularly when run for a sufficiently long time ($i=255$ time steps in the ImageNet experiments), suggests that such an extended duration may not be necessary to generate high-quality images. Instead, we propose making a U-Turn: identifying an appropriate time $T_u$ (and equivalently discrete index $i_u< 255$), and then initializing the reverse process using the score function (SF) as in Eq.~(\ref{eq:reverse_sde}), or its NN- approximation, at this shorter time. The initialization involves a sample generated from the explicitly known distribution $p_t(\cdot|{\bm x}_0) = \mathcal{N}(\cdot|{\bm 0}; \beta(t) \hat{\bm I})$, that is conditioned to a particular choice of a GT sample for ${\bm x}_0$ . See Algorithm \ref{alg:cap}.

\begin{algorithm}[h!]
\caption{U-Turn}\label{alg:cap}
\begin{algorithmic}
\Require ${\bm s}_{\bm \theta}({\bm x}_t,t)$ -- NN approximation of ${\bm s}_{\bm \theta}({\bm x}_t,t)$ defined by Eq.~(\ref{eq:s-exact}); ${\bm x}_0 \sim p_{\text{GT}}$, $T_u$
\begin{enumerate}
    \item Initialize the reversed process: ${\bm y}_{T_u} \sim \mathcal{N}(\cdot|{\bm x}_0; \beta(T_u) \hat{\bm I})$
    \item Run the reversed process according to Eq.~(\ref{eq:reverse_sde}).
    \item Output the newly generated/synthetic image ${\bm y}_0$.
\end{enumerate}
\end{algorithmic}
\end{algorithm}

The algorithm depends on $T_u$, raising the significant question: how do we identify an optimal, or simply appropriate $T_u$? One option, suggested by the analysis in Section \ref{sec:basic-SBD}, is to introduce criteria based on one of the tests --  KS test or SF-norm test. For example, we can choose $T_u$ by setting the SF-norm such that its derivative reaches a predefined small value, indicating that the SF-norm stops changing.

Alternatively, we can experiment by scanning different values of $T_u$. We can start by relying on both our visual perception and FID tests of the U-Turn.

\subsection{Visual Examination on the ImageNet Model}

\begin{figure}[h!]
    \centering
    \begin{tabular}{lc}
        \begin{turn}{90} \quad \tiny GT \end{turn}&
        \includegraphics[scale=0.5]{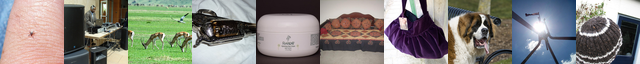}\\
        
        \begin{turn}{90} \quad \tiny $30$ \end{turn}&
        \includegraphics[scale=0.5]{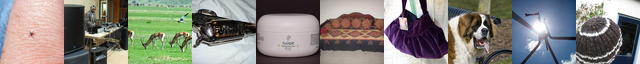}\\
        
        \begin{turn}{90} \quad \tiny $55$ \end{turn}&
        \includegraphics[scale=0.5]{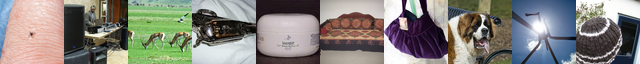}\\
        
        \begin{turn}{90} \quad \tiny $80$ \end{turn}&
        \includegraphics[scale=0.5]{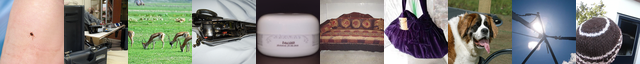}\\
        
        \begin{turn}{90} \quad \tiny $105$ \end{turn}&
        \includegraphics[scale=0.5]{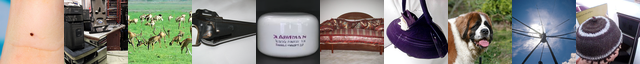}\\
        
        \begin{turn}{90}\quad \tiny $130$ \end{turn}&
        \includegraphics[scale=0.5]{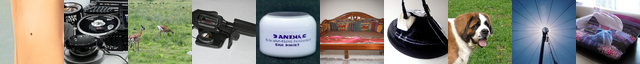}\\
        
        \begin{turn}{90}\quad \tiny $155$ \end{turn}&
        \includegraphics[scale=0.5]{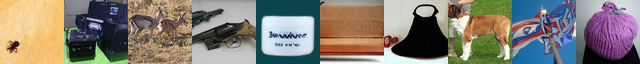}\\
        
        \begin{turn}{90}\quad \tiny $180$ \end{turn}&
        \includegraphics[scale=0.5]{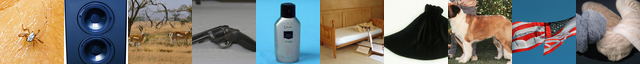}\\
        
        \begin{turn}{90}\quad \tiny $205$ \end{turn}&
        \includegraphics[scale=0.5]{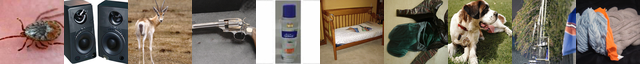}\\
        
        \begin{turn}{90}\quad \tiny $230$ \end{turn}&
        \includegraphics[scale=0.5]{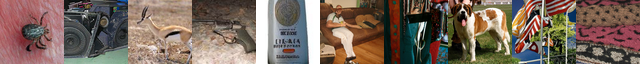}\\
        
        \begin{turn}{90}\quad \tiny $255$ \end{turn}&
        \includegraphics[scale=0.5]{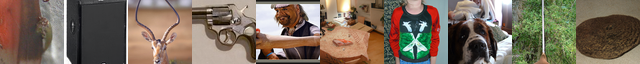}\\

    \end{tabular}

    \caption{ImageNet Visualization: U-Turn at different times $T_u$ with inputs of the forward and reverse processes conditioned to the same class. 
    \label{fig:ImageNet-visual}}

\end{figure}

Images generated using Algorithm \ref{alg:cap} applied to the ImageNet dataset are displayed in Fig.~\ref{fig:ImageNet-visual}. These figures illustrate the evolution of output images in the U-Turn scheme as a function of \( T_u \), conditioned on the class (note that the class remains unchanged from the GT sample through to the initialization of the reverse process after the U-Turn).

A visual inspection reveals a transition around \( i_u \in [150, 200] \), where the model shifts from reproducing the initial images observed at earlier times to generating new images from the same class at later times. We term this dynamic (phase) transition the \textbf{Memorization Transition} and denote the corresponding transition time as \( T_m \). Furthermore, within experiments displayed in the same column of Fig.~\ref{fig:ImageNet-visual} -- each using a different \( T_u \) but initialized with the same image and random seed for the reverse process -- the transition appears relatively sharp. In other words, we consistently observe clear images rather than artifacts (e.g., noisy images or mixed representations). Notably, for \( T_u > T_m \), the generated images vary with different values of \( T_u \) within the same column.

Moreover, moving from one column to the next -- which corresponds to different initial images and labels -- reveals some significant variation in the specific time at which the transition \( T_m \) occurs, that is dependence on the GT sample and its class -- see Section \ref{sec:cond-U-Turn} for further discussion of the strong sensitivity to the GT sample.

In summary, based on our visual analysis of Fig.~\ref{fig:ImageNet-visual}, we conclude that to generate new images, a U-Turn at \( T_u > T_m \) is required. The value of \( T_m \) appears to depend on both the class and the underlying SBD model.

\subsection{FID Test: U-Turn vs Artificial Initialization} \label{sec:U-Turn-VS-Artificial}

\begin{figure}[H]    \subfigure{
        \centering
        \includegraphics[scale=0.4]{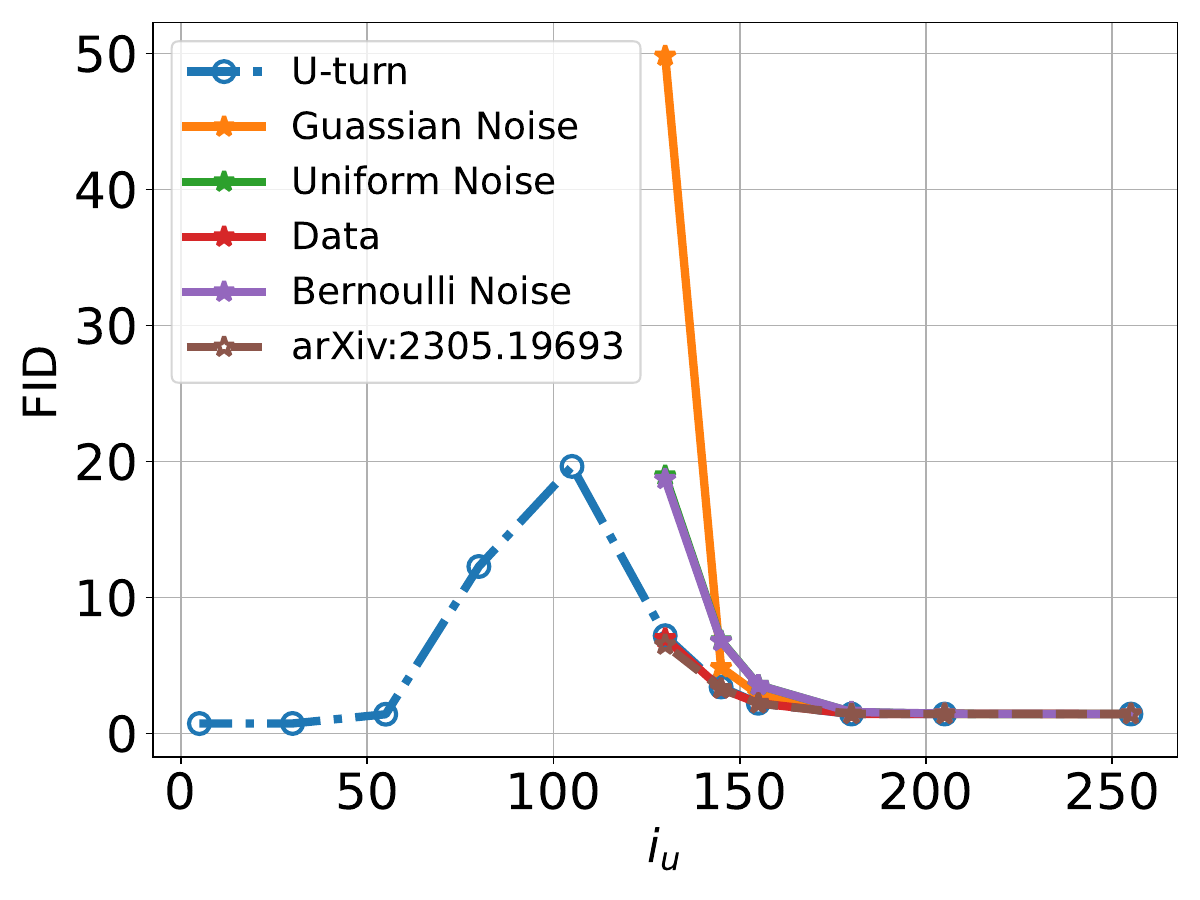} }
    \subfigure{
        \centering
        \includegraphics[scale=0.4]{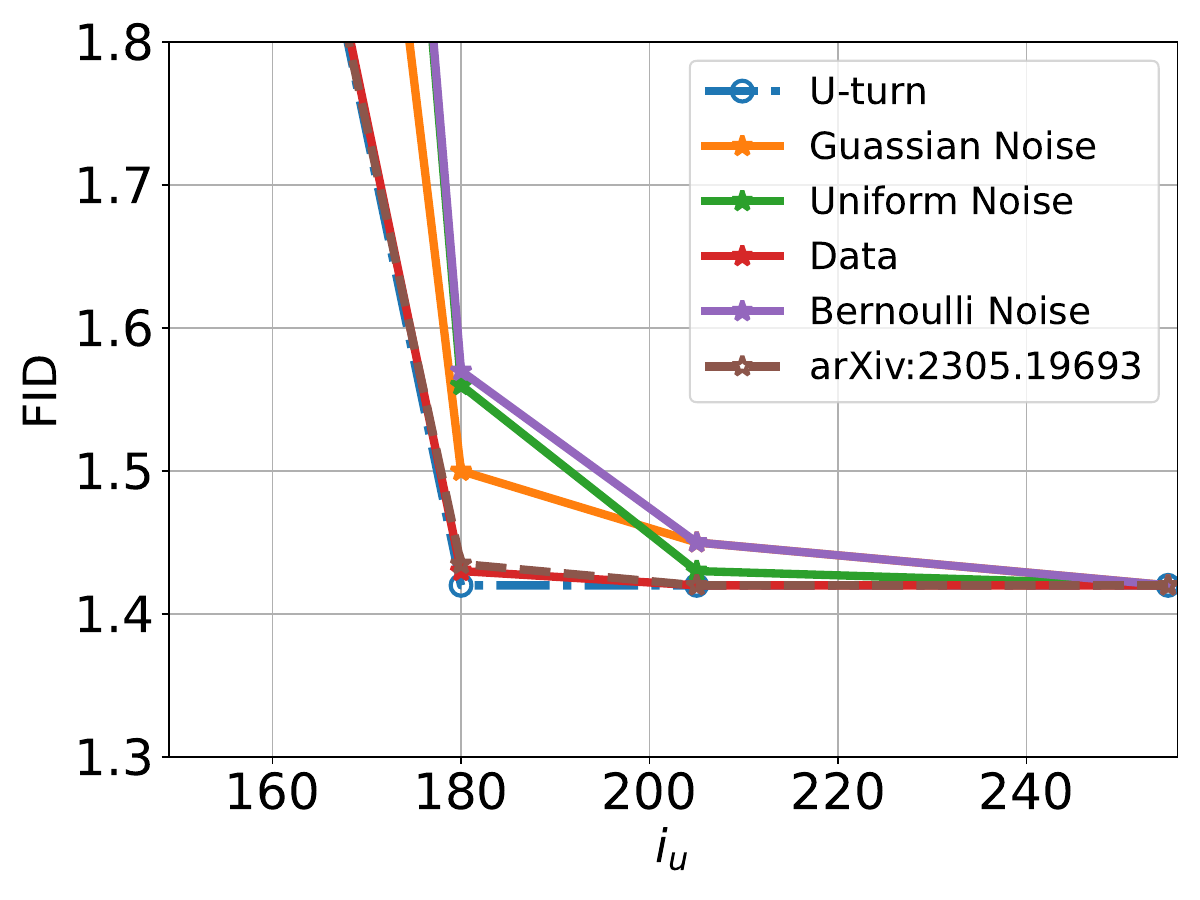} }
    \caption{FID score of U-Turn and its variations with different initializations of the reverse process as a function of $i_u$ (the discrete index of $T_u$). The right sub-figure is a zoomed-in version of the left sub-figure.  
    \label{fig:noise_schedulers}}
\end{figure}

Let's analyze the U-Turn quantitatively using the FID score. However, in addition to running the U-Turn algorithm in its basic form, as shown in Algorithm \ref{alg:cap}, we will experiment with replacing the U-Turn-specific initialization of the reverse process with ${\bm y}_{T_u}$ generated artificially -- independently of the GT sample used in Algorithm \ref{alg:cap}. The results of these experiments are shown in Fig.~(\ref{fig:noise_schedulers}). Several useful observations can be deduced from these results.

First, we observe that a low FID score for the U-Turn alone is not indicative of the diversity of the generated images. When the U-Turn is made at a sufficiently small $T_u$, a low FID simply corresponds to memorization of the initial GT image. Conversely, when we test the FID of the U-Turn jointly with the initialization of the reverse process at the same $T_u$, but with a sample independent of the GT image, these combined results indicate whether respective $T_m$ has been reached. Specifically, if the two processes show comparable FID scores, then $T_u$ is optimal or close to optimal. We also see that if $T_u \geq T_m$, a number of alternative initializations become comparable to the U-Turn in terms of their FID performance. Finally, all artificial initializations result in ridiculously large FID scores if $T_u$ is too short.

We conclude that the "U-Turn vs artificial initialization" comparison provides an explicit method for finding the optimal U-Turn time, $T_m$. For the ImageNet case, this suggests that $i_m \approx 200$ (the index of $T_m$).

\subsection{Auto-Correlation Function of U-Turn}\label{sec:U-AC}

\begin{figure}[H]
  \begin{center}
\includegraphics[scale=0.5]{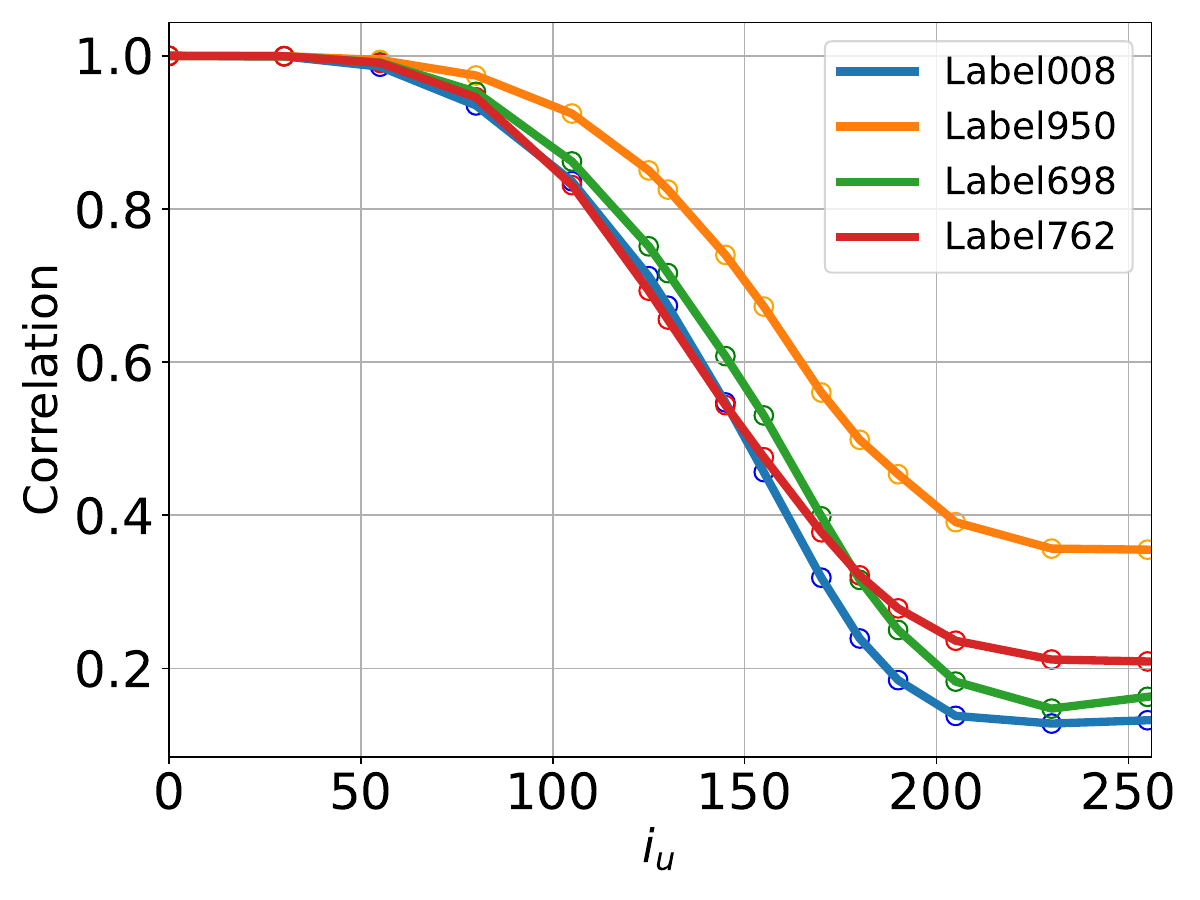}
    \vspace{-20pt}
\caption{ImageNet: U-Turn Auto-Correlation functions for $C_{UT}(T_u)$, defined in Eq.~(\ref{eq:direct-independence}), conditioned to labels and for each label evaluated over 1000 generated samples. See Fig.~(\ref{fig:ks_test_many_class_ImageNet}) for description of the classes.}
\label{fig:ImageNet-UTurn}
  \end{center}
\end{figure}

The approach described in the previous subsection, Section \ref{sec:U-Turn-VS-Artificial}, is strong as it allows for a direct test of the U-Turn performance. This contrasts with the implicit suggestions for selecting $T_u$ based on KS and  SF-norm tests, which track the temporal evolution within the basic SBD algorithm. However, one limitation of the "U-Turn vs Artificial Initialization" test in Section \ref{sec:U-Turn-VS-Artificial} is its empirical nature. We aim to design a test that explicitly refers to the U-Turn and relies on a measure of correlation loss within the U-Turn process between the GT sample and the resulting image. The following construction meets this requirement.

Here, we test the "independence" of the newly generated synthetic samples from their respective GT samples for U-Turns at different $T_u$. Specifically, we compute the U-Turn Auto-Correlation (AC) function:
\begin{gather}\label{eq:direct-independence}
C_{UT}(T_u)=\frac{1}{N}\sum_{n=1}^N \frac{({\bm x}^{(n)}(0))^T {\bm y}^{(n)}(0)}{\left({\bm x}^{(n)}(0)\right)^2},
\end{gather}
where ${\bm x}^{(n)}(0)$ is the $n$-th sample from the GT, resulting in ${\bm x}^{(n)}(T_u) \sim P_{\text{forw. proc.}}(\cdot|T_u;{\bm x}^{(n)}(0))$ generated at $t=T_u$ and then used to initialize the reversed process at the same time, producing a sample path ${\bm y}^{(n)}(T_u \to 0)$ that arrives at $t=0$ as ${\bm y}^{(n)}(0)$. This type of AC function will decay with $T_u$, allowing us to directly quantify when to stop based on a sufficiently small value, which may be dependent on the class.

The results for $C_{UT}(T_u)$ are presented in Fig.~(\ref{fig:ImageNet-UTurn}), which shows a monotonic decrease in $C_{UT}(T_u)$ with increasing $T_u$ ($i_u$) across all classes. We observe that the curves eventually saturate, though at values that vary considerably across classes. We attribute this phenomenon to certain distinctive features that may be relatively uniform within a given class -- for example, the predominance of orange hues within the "oranges (fruits)" class -- which can sustain higher U-Turn auto-correlations within the class even at large $T_u$.

This observation suggests that the most reliable approach for determining $T_m$ might involve setting a threshold based on visual perception of image quality. However, this threshold should be adjusted according to the cross-correlation values within the class, specifically the value of the U-Turn autocorrelation function at the maximum time, $C_{UT}(T=255)$. Based on this, as well as our visual assessment (see Figs.~(\ref{fig:ImageNet-visual})), we propose a practical criterion by setting $1.2 \times C_{UT}(T=255)$ as the threshold, i.e., $C_{UT}(T_m) = C_{UT}(T=255)$. For our test case with ImageNet, this threshold corresponds to approximately $T_m = i_m \approx 200$, though variations are observed across different classes.

\subsubsection{Gaussian Approximation for the U-Turn Auto-Correlation Function}
\label{sec:UT-AC-ImageNet-averaged}

\begin{figure}[H]
	\vspace{-10pt}
	\centering
	\includegraphics[scale=0.5]{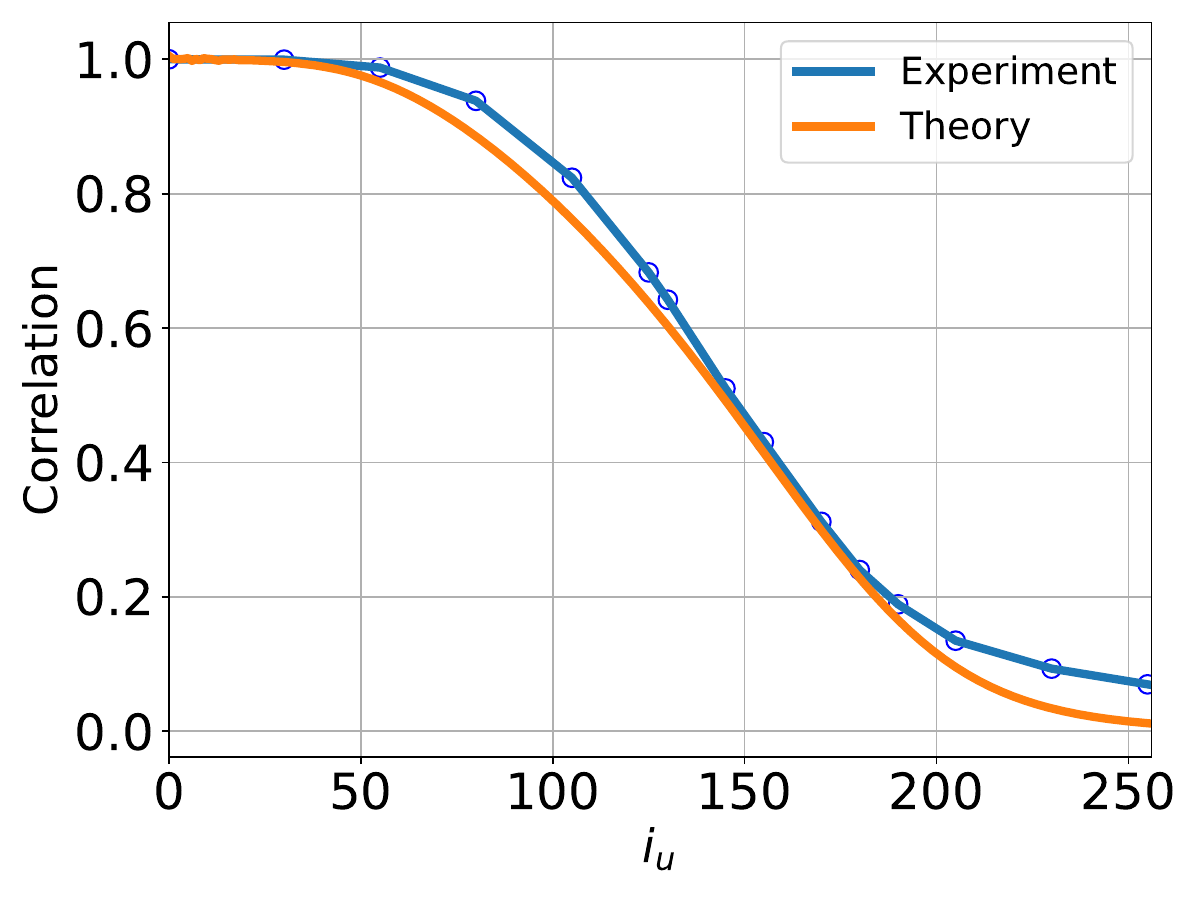}
	\caption{Fully averaged U-Turn auto-correlation functions for the ImageNet dataset, obtained empirically by averaging over all samples and classes, compared to the corresponding fully averaged Gaussian approximation from Appendix~\ref{app:Brownian}.}
	\label{fig:UT-AC-ImageNet}
\end{figure}

If the entire GT dataset were Gaussian, the U-Turn auto-correlation (AC) function could be computed analytically, as both the forward and reverse processes would also be Gaussian in this case. The respective computations are detailed in Appendix \ref{app:Brownian}. Using this Gaussian theory, we approximate the AC function for the GT dataset by computing its covariance while ignoring non-Gaussian contributions.

A comparison between the Gaussian theory and the empirical evaluation of the U-Turn AC function for the entire ImageNet dataset (averaged across all classes) is shown in Fig.~\ref{fig:UT-AC-ImageNet}. Notably, we observe a reasonably good fit between the Gaussian theory and the empirically averaged curve. The close agreement between the two suggests that a global ``annealed'' characterization of the U-Turn behavior is Gaussian, effectively smoothing out sample-specific and class-specific variations which we saw above (in the main part of this subsection) analyzing the U-Turn AC function conditioned to classes.


\section{Multi-Class  -- The Case of CIFAR-10}\label{sec:CIFAR}

We now turn to discussing how the U-Turn approach, along with the performance tests introduced and analyzed thus far, performs in the case of an unsupervised multi-class setting—that is, when a single score function is used to train on datasets containing samples from multiple classes without accounting for the class label of each sample. We begin with a visual inspection of the U-Turn tested on the pre-trained multi-class but single SF CIFAR-10 model.

\subsection{Visual Inspection and FID}\label{sec:CIFAR-visual}

\begin{figure}[h]
    \centering
    \includegraphics[width=\textwidth]{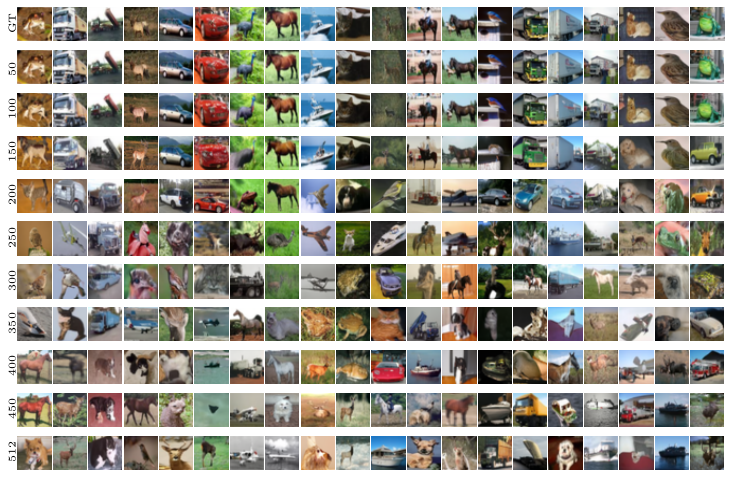}  
    \caption{Unconditional CIFAR-10 image generation. Forward dynamic (computations) is variance exploding (driftless). Reverse dynamics (simulations) is discretized with 512 steps with stochastic samplers.}
    \label{fig:CIFAR-visual}
\end{figure}

Since our primary motivation for experimenting with CIFAR-10 is to gain a better understanding of the multi-class nature of the dataset and its representation within a single (cumulative) SF, we begin our visual analysis of Fig.~(\ref{fig:CIFAR-visual}) by noting that, at sufficiently large values of $T_u$, the U-Turn process starts generating images from different classes. This observation aligns with the predictions of \cite{biroli_dynamical_2024} regarding the emergence of the {\bf Speciation Transition}\footnote{It is worth noting, however, that the findings of \cite{biroli_dynamical_2024} pertain to the standard SBD setting, not specifically our U-Turn modification. Therefore, a more accurate statement would be that we observe a generalization of the Speciation Transition predicted in \cite{biroli_dynamical_2024} to the U-Turn setting.}. We estimate that, depending on the column (each representing experiments with the same initial image and progressively increasing values of $T_u$ from top to bottom), the transition occurs in the range $T_s \in [250,400]$.

Interestingly, in some cases, images from new classes appear early on, but at larger values of $T_u$, the original class can re-emerge. We attribute this to the fact that the probability of the initial state of the reverse process falling within the domain associated with the same label as the GT sample remains non-zero.

Next, we observe that memorization of the initial image ceases to be an issue at somewhat earlier times, which we naturally identify as the {\bf Memorization Transition}, $T_m$, noting that $T_m < T_s$. As in the previously discussed case with a single-class SF, the values of $T_m$ vary from column to column (from one initial GT sample to another) and are observed within the range $T_m \in [150, 250]$.

\begin{figure}
    \centering
    \includegraphics[scale=0.5]{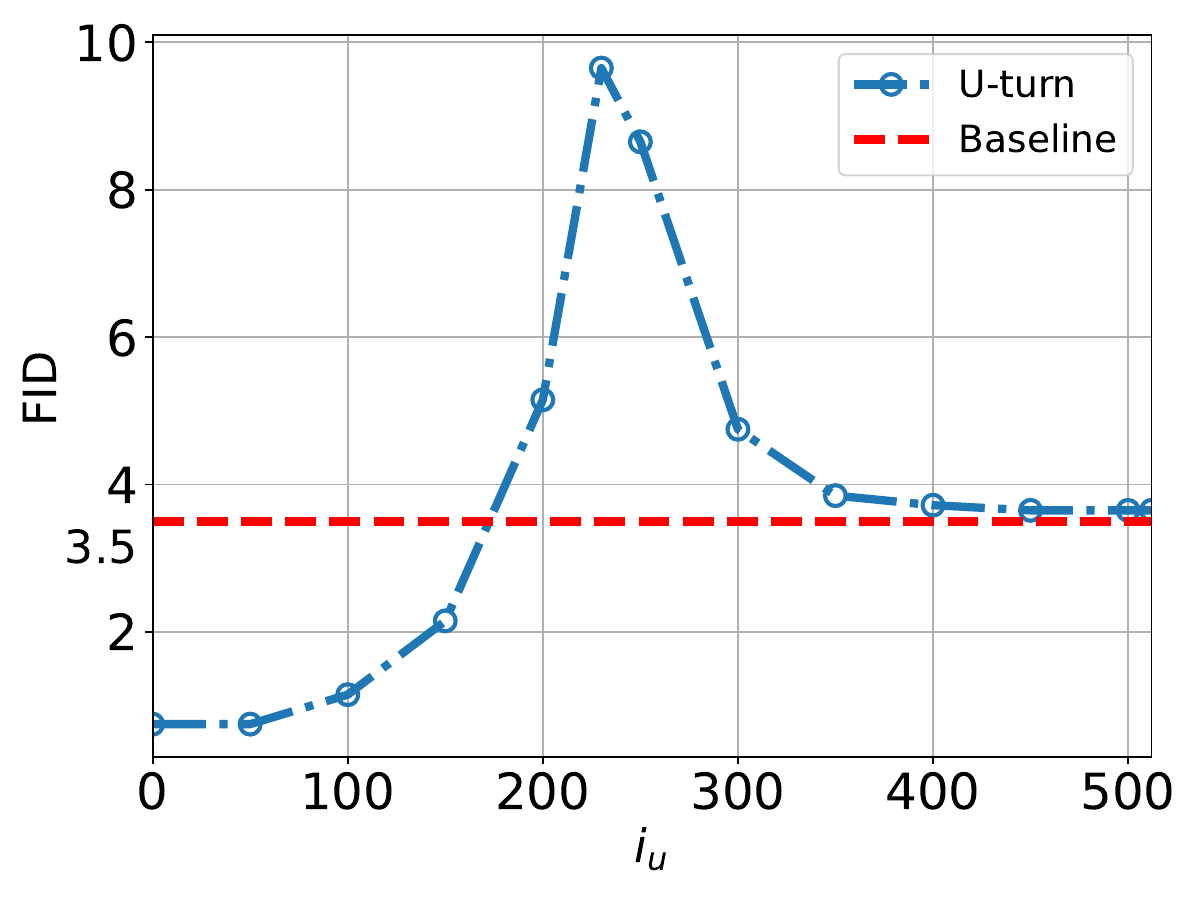}
    \caption{FID score as a function of $T_u$ in U-Turn process applied to CIFAR-10 data. Baseline corresponds to the FID score computed over the data-set for the standard SBD.}
    \label{fig:fid-cifar}
\end{figure}

Turning our attention to image quality (which, as seen in Fig.~(\ref{fig:ImageNet-visual}), remained consistently high across different $T_u$ values in our fixed-class ImageNet experiments), we find that in the multi-class case, image quality varies. In the memorization phase, when $T_u$ is relatively small, image quality is high -- we consistently obtain outputs that are clear and legitimate images. However, as we progress to larger $T_u$ values, beyond the memorization transition but still before the speciation transition $T_s$, we start to observe images within individual sequences (columns in Fig.~(\ref{fig:CIFAR-visual})) that are less clear. Eventually, for sufficiently large values of $T_u > T_s$, the quality of the output images improves again. 

The non-monotonic dependence on $T_u$ of the image quality is also seen quantitatively in Fig.~(\ref{fig:fid-cifar}), where we show FID score computed for CIFAR-10 as a function of $T_u$ in the U-Turn process. 

In summary, our visual examination of Fig.~(\ref{fig:CIFAR-visual}), complemented by the quantitive analysis of the FID in Fig.~(\ref{fig:fid-cifar}), suggest a dynamic phase transition structure within the U-Turn scheme that generally aligns with the observations of \cite{biroli_dynamical_2024} regarding dynamic phase transitions in the standard SBD model: we identify a Memorization Transition at $T_m$ and a Speciation Transition at $T_s$, with $T_m < T_s$. However, we also report new findings specific to the U-Turn approach:
\begin{itemize}
    \item Both $T_m$ and $T_s$ vary depending on the initial GT sample.
    
    \item The region between $T_m$ and $T_s$ appears somewhat blurry, with the potential for some newly generated samples to be noisy or ambiguous (a mix of images).
    
    \item There may be more than one speciation transition --possibly a hierarchy of transitions, each corresponding to diffusion from the domain of the initial GT sample to a domain associated with a different class.
\end{itemize}

\subsection{Conditional U-Turn: Quantitative Analysis}\label{sec:cond-U-Turn}

The concluding remark of the preceding subsection, based on the visual inspection of the newly generated images, should be treated as a hypothesis. Here, we aim to validate this hypothesis through a quantitative analysis, which involves evaluating the results using the U-Turn auto-correlation function introduced in Section~\ref{sec:U-AC}.  

\begin{figure}[H]    \subfigure{
		\centering
		\includegraphics[scale=0.32]{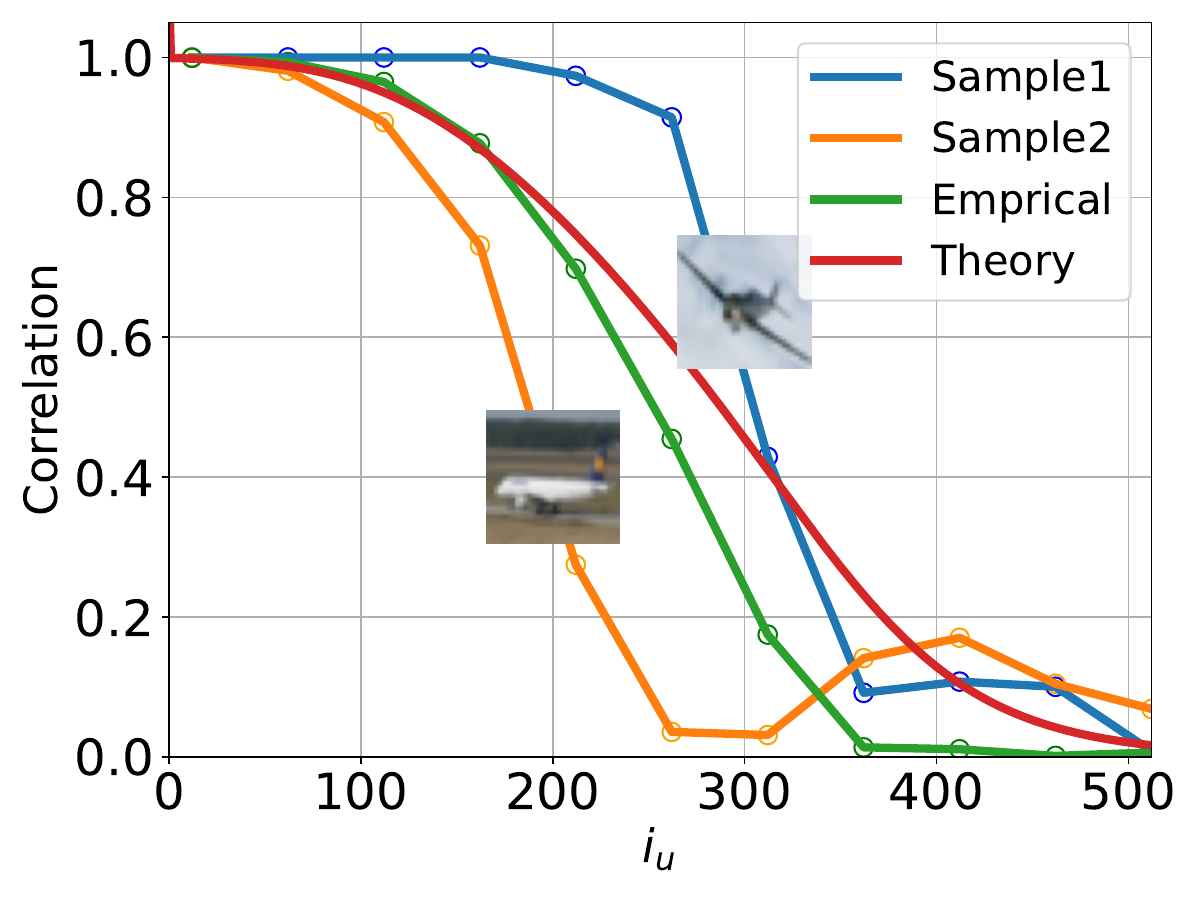} }
	\subfigure{
		\centering
		\includegraphics[scale=0.32]{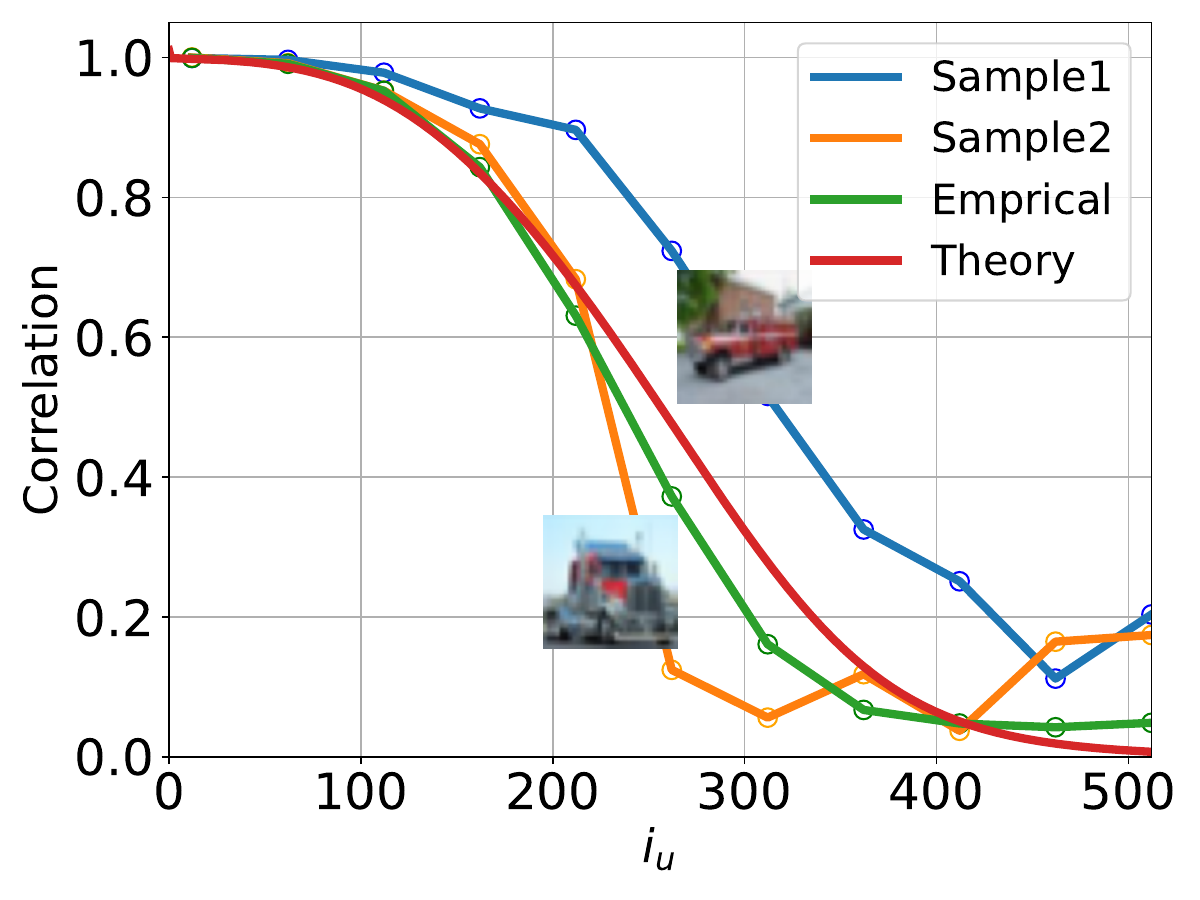} }
        \caption{U-Turn auto-correlation functions conditioned on individual GT samples compared to empirical class-averaged results and the Gaussian approximation from Appendix~\ref{app:Brownian}. \textbf{Left panel:} Results for two distinct GT samples from the ``plane'' class of CIFAR-10. \textbf{Right panel:} Results for two distinct GT samples from the ``truck'' class of CIFAR-10. The analysis highlights the variability of U-Turn correlations across individual samples within the same class, as well as deviations from the Gaussian theoretical prediction.}
		\label{fig:sample_AC}
\end{figure}

Specifically, we select a particular ground truth (GT) image and initialize the U-Turn process using the same image. The resulting trajectories are recorded, and the conditional U-Turn auto-correlation function is computed using Eq.~(\ref{eq:direct-independence}) with \( N = 1 \), corresponding to the selected GT sample. This procedure is repeated for a set of GT samples drawn from various classes, with representative results presented in Fig.~\ref{fig:sample_AC}.

Our analysis of the conditional U-Turn auto-correlations reveals strong dependencies, not only on the class of the initial GT image but also on the specific GT image itself. For example, Fig.~\ref{fig:sample_AC} demonstrates that U-Turn auto-correlations conditioned on different samples -- two GT samples from the ``plane'' class (left panel) and two GT samples from the ``trucks'' class (right panel) -- exhibit noticeable differences both across classes and within samples of the same class.  

In addition, Fig.~\ref{fig:sample_AC} compares these conditional U-Turn auto-correlations with their empirical class-averaged counterparts. The class-averaged curves are obtained by averaging the U-Turn auto-correlation functions over a large number of initial GT samples (1000 samples per class). This comparison highlights the variability induced by individual samples when compared to the smoother, class-averaged behavior.  

Furthermore, we compute the U-Turn correlations under the assumption that the process is Gaussian, utilizing the formulas derived in Appendix~\ref{app:Brownian}. This Gaussian approximation relies primarily on the covariance structure of the GT samples within each class. However, as shown in Fig.~\ref{fig:sample_AC}, the Gaussian approximation fails to closely match the empirical class-averaged results, indicating that the reverse process introduces significant non-Gaussianity. This discrepancy underscores the complexity of the reverse dynamics, which cannot be captured through simple Gaussian assumptions.

In summary, the conditional U-Turn analysis underscores the dual influences of individual GT samples and class-level features on the resulting correlations. Moreover, it reveals substantial non-linearity (and thus non-Gaussianity) of the reverse processes. 


\subsection{From Sample-Conditional to Averaged U-Turn}\label{sec:CIFAR-Tests}

\begin{figure}[H]
\vspace{-10pt}
    \centering
    \includegraphics[scale=0.5]{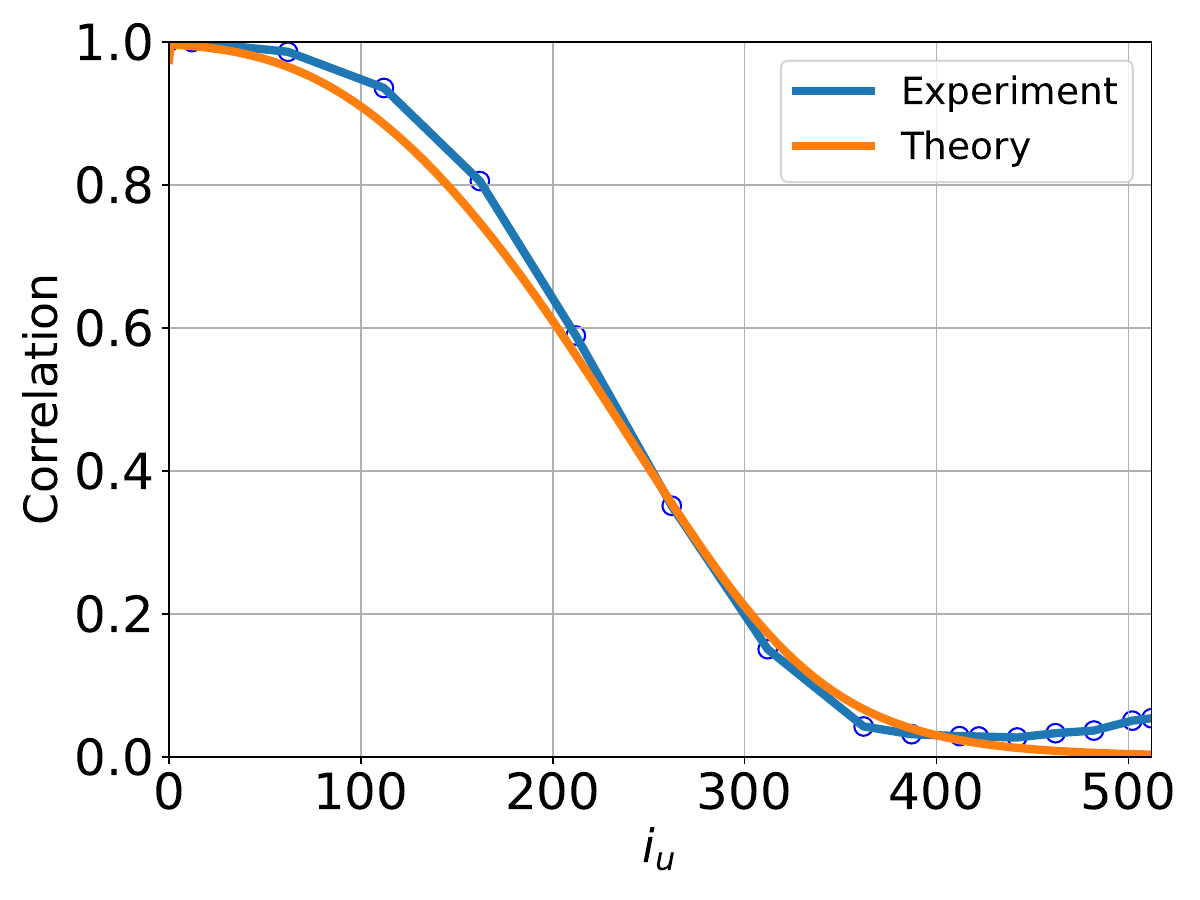}
    \caption{Fully averaged U-Turn auto-correlation functions for the CIFAR-10 dataset, obtained empirically using 5000 samples by averaging over all samples and across all classes, compared to the corresponding fully averaged Gaussian approximation from Appendix~\ref{app:Brownian}. The close agreement between the two provides a global ``annealed'' characterization of the U-Turn behavior, effectively smoothing out sample-specific and class-specific variations.}
    \label{fig:UT-AC-CIFAR}
\end{figure}

We now extend the analysis of the U-Turn auto-correlation function, transitioning from the sample-conditional and class-conditional results to a fully averaged perspective over the CIFAR-10 dataset. In this case, we compute the U-Turn auto-correlation function empirically by averaging over all ground truth (GT) samples and across all classes within CIFAR-10. The corresponding results are shown in Fig.~\ref{fig:UT-AC-CIFAR}.

In this figure, we compare the fully averaged empirical U-Turn auto-correlation function with its Gaussian counterpart derived under the assumptions outlined in Appendix~\ref{app:Brownian}. Unlike the results observed for sample-specific and class-specific conditioning, where significant deviations and non-Gaussian effects were evident, the fully averaged U-Turn auto-correlation function exhibits a good  agreement with the Gaussian approximation.

Recall that we previously reported the results of the U-Turn AC function analysis, averaged over classes, for the ImageNet dataset in Section \ref{sec:UT-AC-ImageNet-averaged}. We now demonstrate that the effective Gaussianity of the class-averaged U-Turn AC function is not specific to the ImageNet dataset but is, in fact, a universal phenomenon, also observed in CIFAR-10.

This observation reveals that while the U-Turn auto-correlation function remains a highly informative object for characterizing the forward and reverse processes, it is not a \emph{self-averaged} quantity in the statistical physics sense. Specifically, our earlier analysis demonstrated considerable variability, nonlinearity, and non-Gaussianity in the conditional U-Turn correlations. However, once averaged across all samples and classes (annealed averaging), these complexities are effectively smoothed out (lost), and the U-Turn behavior becomes almost indistinguishable from that predicted by a Gaussian approximation.

\section{Gaussian Analysis and Gaussian- (G-) Turn}
\label{sec:Gauss}

Appendix \ref{app:Brownian} introduces a Gaussian analysis that models ground-truth (GT) data as Gaussian, extracting the covariance matrix $\hat{\bm \Sigma}_0$ and computing U-Turn correlations under this assumption. This section extends the analysis to explore the non-linearity and non-Gaussianity in empirical processes.

As discussed in Section \ref{sec:cond-U-Turn}, analysis of the conditional U-Turn on CIFAR-10 data reveals that the reversed process is empirically nonlinear and non-Gaussian. Despite this, it is also natural to conjecture that linearity and Gaussianity emerge in the reverse process at least at sufficiently large times $t$ ($T_u>t > T_s$) but may also be approximately correct at the moderate values of $t\in [T_m, T_s]$ starting from the memorization time $T_m$. This means that the U-Turn performed at $T_u > T_s$ and event at $T_u\in [T_m, Ts]$ would therefore render the early stages of the reversed process approximately linear and Gaussian.

Motivated by these insights, we propose the G-Turn procedure, described in Algorithm \ref{alg:g-turn}.

\begin{algorithm}[h!]
\caption{G-Turn}
\label{alg:g-turn}
\begin{algorithmic}
\Require ${\bm s}_{\bm \theta}({\bm x}_t,t)$: NN approximation of ${\bm s}({\bm x}_t,t)$ from Eq.~(\ref{eq:s-exact}); 
$\hat{\bm \Sigma}_0$.
\begin{enumerate}
    \item Initialize a hypothetical reversed process at $T > T_g$ (ideally $T=\infty$) with ${\bm y}_T \sim \mathcal{N}(\cdot|0, \bm{I})$.
    \item Compute (not simulate) ${\bm y}_{T_g}$ using the fully Gaussian procedure described in Appendix \ref{app:Brownian}, treating the forward process as Gaussian at $t=T_g$ with respective mean and variance. Use the time-shifted version of Eq.~(\ref{eq:yt-cond-yT}) for this computation.
    \item Simulate ${\bm y}(t \in [T_g \to 0])$ according to Eq.~(\ref{eq:reverse_sde}), initializing at $t=T_g$ with the result from the previous step. This simulation follows the empirical (nonlinear) score function.
    \item Output the generated synthetic image ${\bm y}_0$.
\end{enumerate}
\end{algorithmic}
\end{algorithm}

\begin{figure}
	\centering
		\includegraphics[width=\textwidth]{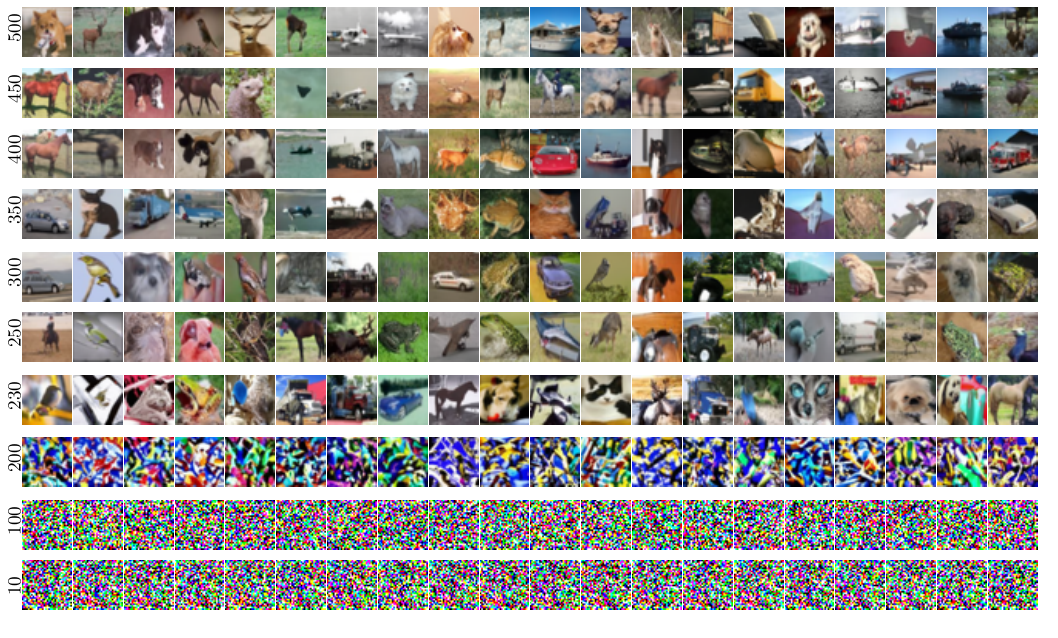}

	\caption{Performance of the G-Turn Algorithm \ref{alg:g-turn} on CIFAR-10. The score function is trained without class conditioning. Columns correspond to different trajectory paths, while rows represent different values of $T_g$.  }
	\label{fig:T_G}
\end{figure}

\begin{figure}
	\centering
		\includegraphics[scale=0.5]{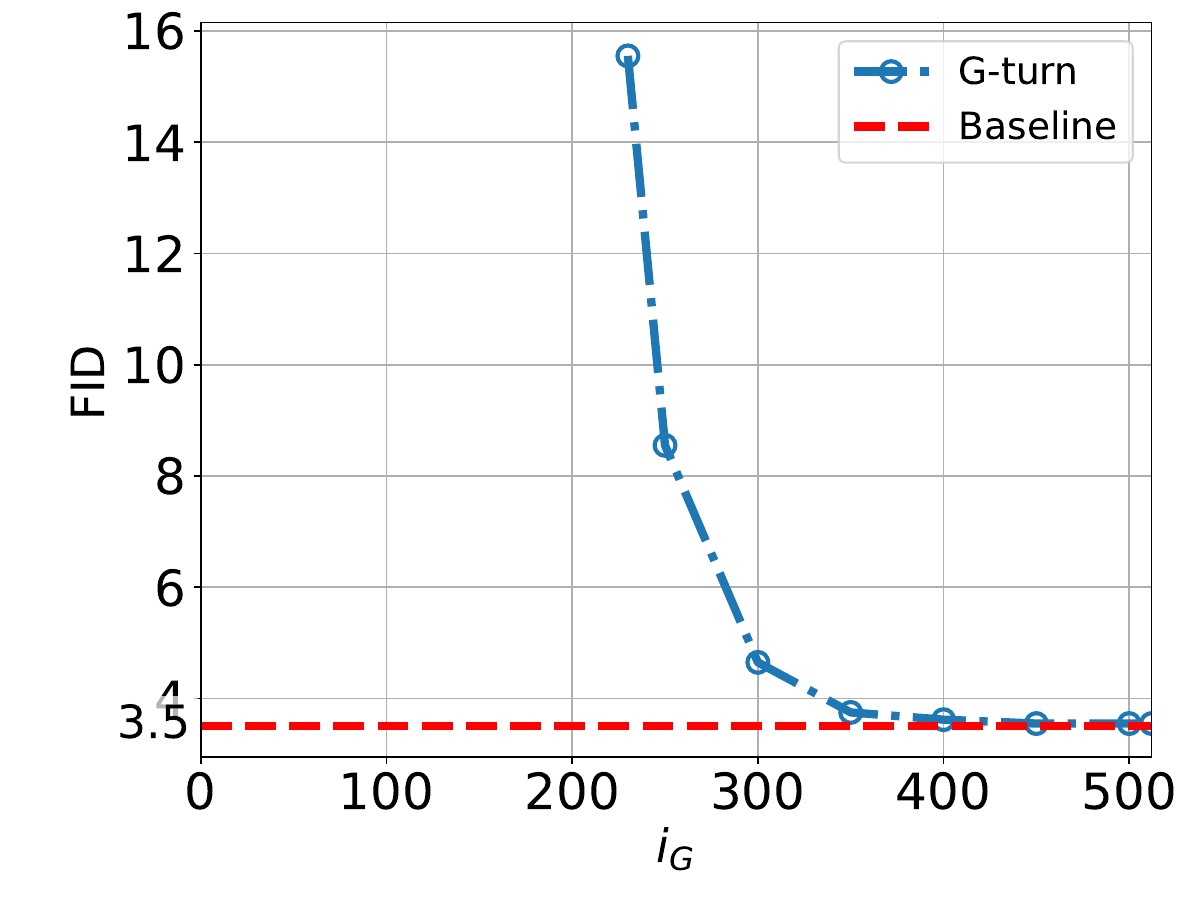}
	\caption{FID scores for the G-Turn Algorithm \ref{alg:g-turn} applied to CIFAR-10, corresponding to the experimental setup visually illustrated in Fig.~\ref{fig:T_G}.}
	\label{fig:FID_G}
\end{figure}

We have applied Algorithm \ref{alg:g-turn} to the unlabeled CIFAR-10 data and the results are shown in Figs.~\ref{fig:T_G},\ref{fig:FID_G}. Visual examination of Fig.~(\ref{fig:T_G}) reveals that the conjecture described above indeed holds true  -- the G-Turn becomes successful only at sufficiently large values of $T_g$ which we estimate to be compatible with the memorization time, $T_s$. However, the quantitative analysis depicted in Fig.~(\ref{fig:FID_G}) also suggests that the FID score of the G-Turn algorithm reaches the lowest value (correspondent to performance of the original SBD algorithm) only when $T_g\approx T_s$. The FID performance is worse (degrades) while still remaining to be reasonably satisfactory when $T_g\in [T_m,T_s]$. 

Note that Algorithm \ref{alg:g-turn} may be compared to the approach described in \cite{raya2023spontaneous}, which initializes the reversed process at $T_g$ with a Gaussian sample which mean and variance are computed empirically. In light of our discussion of Detailed Balance (DB) in Section \ref{sec:stage} (see the footnote), this approach is effectively equivalent to the G-Turn procedure described here.

Also, a related discussion in \cite{wang2024the} addresses the linearization (or Gaussianization) of diffusion samplers. The authors analyze a deterministic reversed process (lacking stochastic components) and develop an analytical model assuming a Gaussian distribution with an arbitrary covariance matrix, represented via singular value decomposition to allow rank deficiency. They provide analytical expressions for the score function and propose approximations based on applying this model to singular value decompositions of actual GT data. A key empirical finding, consistent with ours, is that the linear score function model (Gaussian approximation of the marginal distribution) holds at large times in the forward process (and early times in the reversed process) but fails during the earliest stages of forward evolution.

\section{U-Turn for Deterministic Samplers}\label{sec:deterministic-samplers}

Although the U-Turn is most naturally applied when the diffusion in the reverse process matches that of the forward process—thus ensuring detailed balance, as extensively discussed above—it can also be utilized in scenarios where detailed balance (DB) is broken. Specifically, the U-Turn can be applied to deterministic samplers.  In this case the right hand side in Eq.~(\ref{eq:FP-inverse}) is replaced by zero, the score-function dependent term on the left hand side of the equation is multi-played by $1/2$, and respective corrections are made in Eq.~(\ref{eq:reverse_sde}). 

\begin{figure}[h!]
	\centering
	\begin{tabular}{lc}
		\begin{turn}{90} \ \tiny GT \end{turn}&
		\includegraphics[scale=0.65]{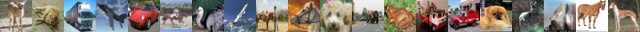}\\
		
		\begin{turn}{90} \  \tiny $4$ \end{turn}&
		\includegraphics[scale=0.65]{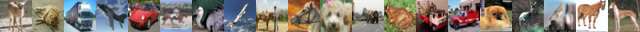}\\
		
		\begin{turn}{90} \ \tiny $6$ \end{turn}&
		\includegraphics[scale=0.65]{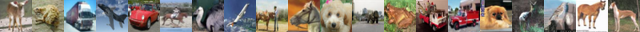}\\
		
		\begin{turn}{90} \ \tiny $8$ \end{turn}&
		\includegraphics[scale=0.65]{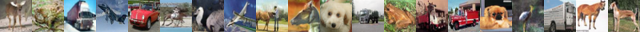}\\
		
		\begin{turn}{90}\ \tiny $10$ \end{turn}&
		\includegraphics[scale=0.65]{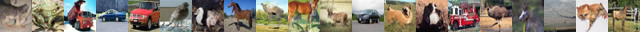}\\
		
		\begin{turn}{90}\ \tiny $12$ \end{turn}&
		\includegraphics[scale=0.65]{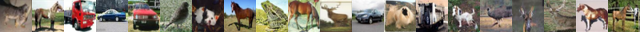}\\
		
		\begin{turn}{90}\ \tiny $14$ \end{turn}&
		\includegraphics[scale=0.65]{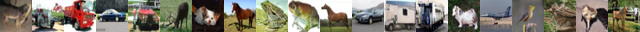}\\
		
		\begin{turn}{90}\ \tiny $15$ \end{turn}&
		\includegraphics[scale=0.65]{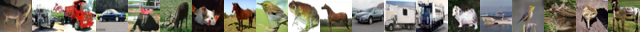}\\
		
		\begin{turn}{90}\ \tiny $16$ \end{turn}&
		\includegraphics[scale=0.65]{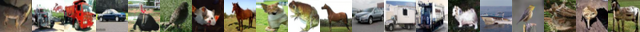}\\
		
		\begin{turn}{90}\ \tiny $17$ \end{turn}&
		\includegraphics[scale=0.65]{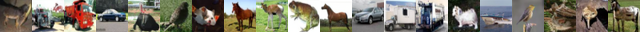}\\
		
		\begin{turn}{90}\ \tiny $18$ \end{turn}&
		\includegraphics[scale=0.65]{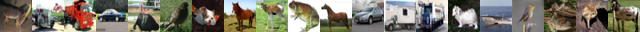}\\
		
	\end{tabular}
	
	\caption{Visual illustration of deterministic sampler for the U-Turn at different times $T_u$ over the unconditional (over classes) CIFAR-10 dataset \cite{karras_elucidating_2022}.}
	\label{fig:det-CIFAR-10}
	
\end{figure}

\begin{figure}[h!]
	\centering
	\begin{tabular}{lc}
		\begin{turn}{90} \quad \tiny GT \end{turn}&
		\includegraphics[scale=0.5]{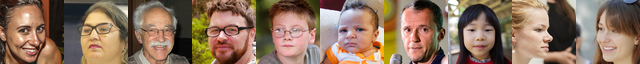}\\
		
		\begin{turn}{90} \quad \tiny $5$ \end{turn}&
		\includegraphics[scale=0.5]{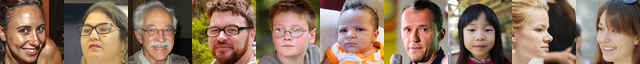}\\
		
		\begin{turn}{90} \quad \tiny $10$ \end{turn}&
		\includegraphics[scale=0.5]{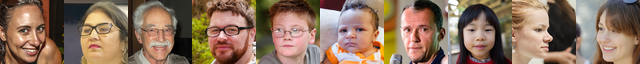}\\
		
		\begin{turn}{90} \quad \tiny $15$ \end{turn}&
		\includegraphics[scale=0.5]{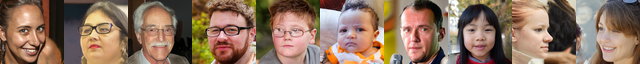}\\
		
		\begin{turn}{90} \quad \tiny $20$ \end{turn}&
		\includegraphics[scale=0.5]{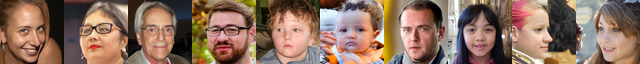}\\
		
		\begin{turn}{90}\quad \tiny $25$ \end{turn}&
		\includegraphics[scale=0.5]{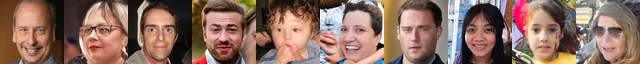}\\
		
		\begin{turn}{90}\quad \tiny $30$ \end{turn}&
		\includegraphics[scale=0.5]{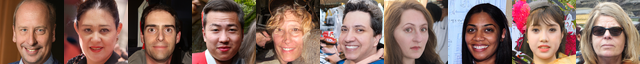}\\
		
		\begin{turn}{90}\quad \tiny $35$ \end{turn}&
		\includegraphics[scale=0.5]{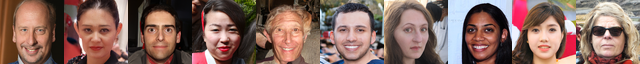}\\
		
		\begin{turn}{90}\quad \tiny $40$ \end{turn}&
		\includegraphics[scale=0.5]{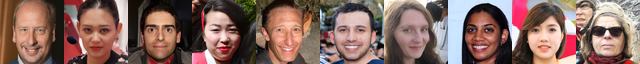}\\	
		
	\end{tabular}
	
	\caption{Visual illustration of the U-Turn at different \( T_u \) in the scheme using the deterministic reverse process (sampler) is presented for the Flickr-Faces-HQ (FFHQ) dataset \cite{karras_elucidating_2022}. This dataset consists of 70,000 high-quality PNG images at 1024×1024 resolution and showcases significant variation in terms of age, ethnicity, and image backgrounds. Additionally, the dataset includes a diverse range of accessories such as eyeglasses, sunglasses, hats, and more.
    } \label{fig:FFHQ}

\end{figure}

We illustrate the performance of the U-Turn for deterministic samplers in Fig.~\ref{fig:det-CIFAR-10} and Fig.~\ref{fig:FFHQ}, applied respectively to models trained unconditionally (across multiple classes) on CIFAR-10 data and Flickr-Faces-HQ (FFHQ) data \footnote{https://github.com/NVlabs/ffhq-dataset}.  

Several observations can be made based on these results:  
\begin{enumerate}
\item {\it Lack of Randomness in the Output Images}: 
   Examining the deterministic sampler for CIFAR-10 (Fig.~\ref{fig:det-CIFAR-10}) and FFHQ (Fig.~\ref{fig:FFHQ}), we observe that when the U-Turn occurs at \( T_u > T_m \), there is no randomness (uncertainty) in the output image. For deterministic reverse processes, the resulting image remains unchanged with further increases in \( T_u \) (at least for sufficiently large \( T_u \)). This is in sharp contrast to stochastic reverse processes, which are the primary focus of this paper, where images generated for different values of \( T_u \) exhibit noticeable variation.  

\item {\it Absence of Speciation Transitions in Some Cases}:  
   In some instances, no speciation transition is observed (e.g., in the case of a car, as shown in the third column of Fig.~\ref{fig:det-CIFAR-10}). This suggests that speciation transitions may not always manifest in individual dynamics but could emerge in some form of averaged behavior.  

\item {\it Intermediate Changes}:  
   In several cases (though not all), there is a range of U-Turn times during which changes in the output image are observed, despite the reverse dynamics being deterministic.  

\item {\it Class Changes with Increasing U-Turn Time}:  
   In some instances, we observe that classes change multiple times as \( T_u \) increases, potentially due to spontaneous and dynamic symmetry breaking.
\end{enumerate}

These observations highlight unique aspects of applying the U-Turn to deterministic samplers and contrast them with the stochastic processes discussed earlier in the paper.

\section{Conclusions and Path Forward}\label{sec:conclusions}

In this work, we proposed the U-Turn diffusion model, an augmentation of the pre-trained Score-Based Diffusion (SBD) model, designed to enhance the efficiency and versatility of synthetic data generation. Our analysis introduced a systematic framework for identifying critical time scales -- Memorization Time (\(T_m\)) and Speciation Time (\(T_s\))  \cite{biroli_dynamical_2024} -- which govern the dynamics of the forward and reverse processes.

Key findings of this study include:
\begin{itemize}
    \item The U-Turn diffusion model significantly reduces the duration of forward and reverse processes while preserving the essential characteristics of the generated samples. This is achieved by initializing the reverse process from a GT sample at \(T_u > T_m\), ensuring adherence to the detailed balance condition.
    
    \item Through a combination of visual and quantitative evaluations—including the Fréchet Inception Distance (FID), Kolmogorov-Smirnov Gaussianity test, Score Function norm test, and U-Turn Auto-Correlation (AC) function test -- we demonstrated the robustness of the U-Turn approach. These analyses revealed that \(T_m\) marks the point where generated samples diverge from the ground truth (GT) samples, while \(T_s\) represents the onset of speciation into distinct classes.

    \item Experiments on ImageNet and CIFAR-10 datasets confirmed the universality of the U-Turn approach across class-conditional and multi-class settings. Notably, the Speciation Transition (\(T_s\)) was observed as a distinctive phase where samples not only diverge from GT samples but also begin to represent different classes.

    \item The U-Turn experiments highlighted a strong sensitivity of the transition times, \(T_m\) and \(T_s\), to the initial GT sample. This lack of self-averaging indicates that transition times are inherently GT sample-dependent.

    \item To further explore the nature of the reverse process (non)-Gaussianity, we designed the G-Turn method, an alternative to the U-Turn. The G-Turn begins at large times, assuming Gaussian behavior governed by the empirical covariance of the forward process, and proceeds with Gaussian computations until \(T_G\). It then switches to fully non-linear score-function-based simulations for \(t \in [T_G \to 0]\). Our results showed that the G-Turn produces high-quality synthetic samples for \(T_G > T_s\) and acceptable quality for \(T_G \in [T_m, T_s]\), but fails for \(T_G < T_m\). These findings suggest that the score function is affine for \(t > T_s\), approximately linear for \(t \in [T_m, T_s]\), and strongly non-linear for \(t < T_m\).

    \item Application of the U-Turn to deterministic samplers illustrated the flexibility of the approach. It revealed unique dynamics, including a lack of randomness in the output images and the absence of speciation transitions in certain scenarios, distinguishing deterministic samplers from stochastic processes.
\end{itemize}

Our findings highlight the potential of the U-Turn diffusion model to enhance the efficiency of generative AI systems while preserving the quality and diversity of outputs. This work lays the groundwork for future explorations into more efficient and versatile generative frameworks. Below, we provide an incomplete list of future research directions that the authors plan to explore:
\begin{itemize}
    \item While this study utilized the simplest forward process (with the score function pre-trained in prior studies), it may be advantageous to further optimize the forward process by tailoring it to the specific characteristics of the GT data.

    \item We hypothesize that the U-Turn approach can be extended to automatically discover classes in a multi-class dataset when using a single score function (averaged over classes, as in the CIFAR-10 model).

    \item In this work, we have leveraged the critical observation that the NN approximation of the score function within a short initial sub-range of the forward process is pivotal for generalization beyond the GT dataset. A systematic investigation of the generalization-memorization trade-off will be essential for deeper insights.

    \item Assuming the availability of multiple pre-trained score functions for the same dataset (or the same application), it would be of interest to design a hybrid reverse process that combines these functions to achieve performance superior to any individual base model.
\end{itemize}

\section*{Computational Resources}

All experiments were conducted on an A100 80GB GPU provided by Jetstream2. We acknowledge the use of ACCESS/NSF cloud computing resources through the Discover grant "Innovation in Generative Diffusion Modeling" (MTH24000). This grant enabled our exploration of cutting-edge techniques in generative diffusion modeling, facilitated by Jetstream2's high-performance computing capabilities.

\section*{Acknowledgments}

We sincerely thank Giulio Biroli and Marc Mézard for engaging in many stimulating discussions on diffusion models and related approaches, which followed the publication of the first version of this report on arXiv.


\newpage
\appendix
\setcounter{figure}{0}

\section{Quality Metrics}

\subsection{The Fréchet Inception Distance}\label{sec:FID}

The Fréchet Inception Distance (FID) is a widely used metric for evaluating the performance of generative models, especially in image generation. It measures the similarity between generated images and real images by comparing their feature representations obtained from a pre-trained Inception network.

Let $\bm{x} \in \mathbb{R}^n$ be a feature vector obtained from the final pooling layer of the Inception network for a given image. We assume that the feature vectors of real and generated images follow multivariate Gaussian distributions $\mathcal{N}(\boldsymbol{\mu}_r, \hat{\bm \Sigma}_r)$ and $\mathcal{N}(\boldsymbol{\mu}_g, \hat{\bm \Sigma}_g)$, respectively.

The FID between the GT and generated/synthetic distributions is defined as:
\begin{equation*}
    \text{FID} = \left\Vert \bm{\mu}_{GT} - \bm{\mu}_g \right\Vert_2^2 + \text{Tr}\left(\hat{\bm \Sigma}_{GT} + \hat{\bm \Sigma}_g - 2\left(\hat{\bm \Sigma}_{GT}^{1/2}\hat{\bm \Sigma}_g\hat{\bm \Sigma}_r^{1/2}\right)^{1/2}\right)
\end{equation*}
where $\bm{\mu}_{GT}$ and $\hat{\bm \Sigma}_{GT}$ are the mean and covariance of the GT image features, $\bm{\mu}_g$ and $\hat{\bm \Sigma}_g$ are the mean and covariance of the generated/synthetic image features, and $\text{Tr}(\cdot)$ denotes the trace of a matrix.

\subsection{Kolmogorov-Smirnov Test}\label{sec:KS}

The Kolmogorov-Smirnov (KS) test is a non-parametric test used to determine if a sample comes from a specific distribution. We define the KS-ratio as the number of dimensions $k$ for which the Gaussianity of the corresponding $x_k(t)$ is not confirmed, divided by the total number of dimensions (the cardinality of $\bm{x}(t)$, which is $64\times 64\times 3$ in our numerical experiments):
\[
\text{KS}(t) = \frac{\text{\# of } p_t(x_k) \text{ failing the Gaussianity test}}{\text{cardinality of } \bm{x}}.
\]

We opted for the KS test over the Kullback-Leibler (KL) test for several reasons. Firstly, KL divergence is asymmetric and not a true distance metric, while the KS test is symmetric, assessing the maximum separation between the test and reference distributions. Secondly, the KS test is non-parametric, meaning it doesn't depend on the parameters of the benchmark distribution, unlike KL divergence. Thirdly, the KS test is computationally simpler and more suited for our analysis, which does not focus on distributions with non-trivial tails.

The one-sample KS test evaluates the hypothesis that a sample $X_1, X_2, \dots, X_n$ comes from a specified probability distribution $F(x)$. The test statistic is the maximum distance between the empirical distribution function $F_n(x)$ and the hypothesized cumulative distribution function $F(x)$:
\[
D_n = \sup_x |F_n(x) - F(x)|.
\]
The null hypothesis $H_0$ is that the sample comes from the specified distribution $F(x)$. The null hypothesis is rejected at a given significance level $\alpha$ if $D_n$ exceeds the critical value $D_\alpha$.

The critical values $D_\alpha$ for the one-sample KS test depend on the sample size $n$ and the chosen significance level $\alpha$. The significance level $\alpha$ represents the probability of rejecting the null hypothesis when it is true. Common choices for $\alpha$ are:

\begin{itemize}
    \item $\alpha = 0.01$ (1\% significance level)
    \item $\alpha = 0.05$ (5\% significance level)
    \item $\alpha = 0.10$ (10\% significance level)
\end{itemize}

For a given sample size $n$ and significance level $\alpha$, the critical value $D_\alpha$ can be obtained from statistical tables or calculated using statistical software. Typical critical values for different sample sizes and significance levels are shown in Table \ref{tab:ks_critical_values}.

\begin{table}[h]
    \centering
    \caption{Critical Values for the One-Sample Kolmogorov-Smirnov Test}
    \label{tab:ks_critical_values}
    \begin{tabular}{cccc}
        \hline
        Sample Size $n$ & $\alpha = 0.01$ & $\alpha = 0.05$ & $\alpha = 0.10$ \\
        \hline
        10 & 0.368 & 0.410 & 0.490 \\
        50 & 0.230 & 0.192 & 0.172 \\
        100 & 0.122 & 0.136 & 0.163 \\
        1000 & 0.038 & 0.043 & 0.052 \\
        \hline
    \end{tabular}
\end{table}

If the test statistic $D_n$ is greater than the critical value $D_\alpha$ for the chosen significance level $\alpha$, the null hypothesis is rejected, indicating that the sample does not come from the hypothesized distribution $F(x)$. Otherwise, if $D_n \leq D_\alpha$, the null hypothesis cannot be rejected at the chosen significance level \cite{ks1951}.

\subsection{Average of the Score Function 2-Norm} \label{sec:score-norm}
The normalized average of the score function 2-norm, denoted by $S(t)$, provides a way to analyze the evolution of the score function's magnitude over time, relative to its initial state. By taking the expected value of the squared 2-norm of this score function and normalizing it by the expected value of the squared 2-norm at the initial time $t=0$, we obtain a dimensionless quantity $S(t)$. This normalization allows us to track the relative change in the score function's magnitude as the system evolves, providing insights into the dynamics of the underlying probability distribution.

\begin{equation}
	S(t)\doteq \sqrt{\frac{\mathbb{E}\left[\left(\nabla_{\bm x} \log p_t({\bm x})\right)^2 \right] }{\mathbb{E}\left[\left(\nabla_{\bm x} \log p_0({\bm x})\right)^2\right]}},
	\label{eq:normalized_average-score}
\end{equation}

\section{Brownian Diffusion}\label{app:Brownian}

In this Appendix, we provide detailed calculations for an SBD model where the forward process is modeled as Brownian diffusion -- specifically, a model with zero drift and a time -- evolving, space-independent variance. 

Consider the case of a time-dependent Brownian motion where the forward dynamics 
\begin{gather}\label{eq:xt-br}
d\bm{x}_t = \sqrt{2\beta_t} d\bm{w}_t, \quad \bm{x}_0 \sim \mathcal{N}\left({\bm 0}, \hat{\bm \Sigma}_0\right),
\end{gather}
has no drift, therefore leading to a stochastic process with increasing variance, $t\geq t'\geq 0$:
\begin{gather}\label{eq:xt-br-cond}
\bm{x}_t\  \big|\ {\bm x}_{t'} \sim   \mathcal{N}\left({\bm x}_{t'}, 2 \bm{I} \int_{t'}^t ds \beta_s\right),\  \bm{x}_t\sim \mathcal{N}\left({\bm 0}, \hat{\bm \Sigma}_t\right),\quad \hat{\bm{\Sigma}}_t \doteq  \hat{\bm{\Sigma}}_0 + 2 \bm{I} \int_0^t ds \beta_s.
\end{gather}

Differentiating the dynamic variance and its inverse we arrive at the following equations for the differentials 
\begin{gather}\label{eq:br-sigma-diff}
d\hat{\bm{\Sigma}}_t = 2 \bm{I} \beta_t dt,\quad d\hat{\bm{\Sigma}}^{-1}_t = -2 \beta_t  \hat{\bm{\Sigma}}^{-2}_t dt. 
\end{gather}

The reverse process is governed by 
\begin{gather*}
    d\bm{y}_t = 2\beta_t  \hat{\bm \Sigma}^{-1}_{t} \ \bm{y}_t  dt + \sqrt{2\beta_t } d\bar{\bm w}_t,\quad {\bm y}_{T_u}={\bm x}_{T_u},
\end{gather*}
where we accounted for the Gaussianity of the marginal probability distribution (of both forward and reverse processes -- which are assumed equal by the definition of how we constructed the reverse process),  ${\bm \nabla}_{\bm{x}_t} \log p_{t}(\bm{x}_t) = - \hat{\bm \Sigma}^{-1}_t \bm{x}_t$.

Next,  let us introduce the auxiliary dynamic variable, $\bm{z}_t = \hat{\bm \Sigma}^{-1}_{t} \bm{y}_t$ \cite{pierret2024}, and utilizing Eq.~(\ref{eq:br-sigma-diff}) we arrive at 
\begin{align*} 
	d\bm{z}_t & =  d(\hat{\bm \Sigma}^{-1}_{t})  \bm{y}_t  + \hat{\bm \Sigma}^{-1}_{t} d\bm{y}_t \\ 
 & = -2\beta_t \hat{\bm \Sigma}^{-2}_{t} \bm{y}_t dt + \hat{\bm \Sigma}^{-1}_{t} (2\beta_t  \hat{\bm \Sigma}^{-1}_{t} \ \bm{y}_t  dt + \sqrt{2\beta_t } d\bar{\bm w}_t) = \sqrt{2\beta_t } \hat{\bm \Sigma}^{-1}_{t}  d\bar{\bm w}_t,
\end{align*}
then resulting in 
\begin{align}\label{eq:yt-br-f}
\bm{y}_t  & =  \hat{\bm \Sigma}_{t} \hat{\bm \Sigma}^{-1}_{T_u} \bm{y}_{T_u} +  \hat{\bm \Sigma}_{t} \int_{T_u}^t \sqrt{2\beta_s}\  \hat{\bm \Sigma}^{-1}_{s} \ d\bar{\bm w}_s\\ \nonumber & = \hat{\bm \Sigma}_{t} \hat{\bm \Sigma}^{-1}_{T_u} \left({\bm x}_0+\int_0^{T_u}\sqrt{2\beta_s}d{\bm w}_s\right)+  \hat{\bm \Sigma}_{t} \int_{T_u}^t \sqrt{2\beta_s}\  \hat{\bm \Sigma}^{-1}_{s} \ d\bar{\bm w}_s,
\end{align}
where first and second lines express conditioning to $\bm{y}_{T_u}$ and to ${\bm x}_0$, respectively. Averaging Eq.~(\ref{eq:yt-br-f}) over the independent forward, ${\bm w}_t$, and reverse, $\bar{\bm w}_t$, Wiener processes we derive:
\begin{align}\nonumber 
    {\bm y}_t \ \big|\ \bm{y}_{T_u} & \sim {\cal N}\left(\hat{\bm \Sigma}_{t} \hat{\bm \Sigma}^{-1}_{T_u} \bm{y}_{T_u},\hat{\bm \Omega}_{t;r}\right),\quad \hat{\bm \Omega}_{t;r} \doteq  
\text{Var}\left( \hat{\bm \Sigma}_{t} \int_{t}^{T_u} \sqrt{2\beta_s}\  \hat{\bm \Sigma}^{-1}_{s} \ d\bar{\bm w}_s \right)\\ \label{eq:yt-cond-yT} & =  \hat{\bm \Sigma}_{t} \int^{T_u}_t 2\beta_s \hat{\bm \Sigma}^{-2}_{s} ds \hat{\bm \Sigma}_{t}= -\hat{\bm \Sigma}_{t} \hat{\bm \Sigma}^{-1}_{s}|^{T_u}_t \hat{\bm \Sigma}_{t}
= \hat{\bm \Sigma}_{t} -\hat{\bm \Sigma}_{t}  \hat{\bm \Sigma}^{-1}_{T_u}\hat{\bm \Sigma}_{t},\\
    \nonumber 
    {\bm y}_t \ \big| \ {\bm x}_0 & \sim {\cal N}\left(\hat{\bm \Sigma}_{t} \hat{\bm \Sigma}^{-1}_{T_u}{\bm x}_0,  \hat{\bm \Sigma}_{t} -\hat{\bm \Sigma}_{t}  \hat{\bm \Sigma}^{-1}_{T_u}\hat{\bm \Sigma}_{0}\hat{\bm \Sigma}^{-1}_{T_u}\hat{\bm \Sigma}_{t}
    \right). 
\end{align}
The last formula results in the following expression for the U-Turn auto-correlation function:
\begin{gather}\label{eq:U-T-AC-Gauss}
    C_{UT}(T_u)= \frac{\mathbb{E}_{{\bm x}_0}\left[{\bm y}_0^T{\bm x}_0\right]}{\mathbb{E}_{{\bm x}_0}\left[{\bm x}_0^T{\bm x}_0\right]}=\frac{\text{tr}\left(\hat{\bm \Sigma}_{0}^2 \hat{\bm \Sigma}^{-1}_{T_u}\right)}{\text{tr}\left(\hat{\bm \Sigma}_{0}\right)}.
\end{gather}

Combining Eq.~(\ref{eq:yt-cond-yT}) and Eq.~(\ref{eq:xt-br}) we derive the following expression for statistics of ${\bm y}_0$ (not conditioned on ${\bm x}_0$):
\begin{gather}\label{eq:x0-stat-uncond}
    {\bm y}_0\sim {\cal N}\left({\bm 0},\hat{\bm \Sigma}_0\right),
\end{gather}
which confirms what we expect (according to the Anderson theory) that statistics of ${\bm y}_0$ is exactly equal to statistics of ${\bm x}_0$ at all values of $T_0$. 

We can also extend Eq.~(\ref{eq:yt-cond-yT}) to quantify correlations between ${\bm y}$ observed at any two different times within the reversed process:
\begin{gather}\label{eq:yt-cond}
    0\leq t'\leq t:\quad {\bm y}_{t'} \ \big|\ \bm{y}_{t} \sim {\cal N}\left(\hat{\bm \Sigma}_{t'} \hat{\bm \Sigma}^{-1}_{t} \bm{y}_{t},\hat{\bm \Sigma}_{t'} -\hat{\bm \Sigma}_{t'}  \hat{\bm \Sigma}^{-1}_{t} \hat{\bm \Sigma}_{t'}\right).
\end{gather}

Note, that contrasting Eq.~(\ref{eq:yt-cond}) for the transition probability  from $t$ to $t'$ in the reversed process to Eq.~(\ref{eq:xt-br-cond}) for the transition probability from $t'$ to $t$ in the forward process, and taking into account that the marginal probabilities at $t$ and $t'$ (conditioned to ${\bm x}_0$) are ${\cal N}({\bm x}_0,\hat{\bm \Sigma}_t)$ and ${\cal N}({\bm x}_0,\hat{\bm \Sigma}_{t'})$ respectively, we confirm for our (admittedly very special example) that the Detailed Balance,
\begin{gather}\label{eq:DB}
    p^{(f)}\left({\bm x}_t|{\bm x}_{t'}\right)p({\bm x}_{t'})=p^{(r)}\left({\bm x}_{t'}|{\bm x}_t\right)p({\bm x}_{t}),
\end{gather}
proven in the general case in \cite{anderson_reverse-time_1982}, holds. (Here in Eq.~(\ref{eq:DB}) $p^{(r)}\left(\cdot|\cdot\right)$ and $p^{(f)}\left(\cdot|\cdot\right)$ are notations for the transition probabilities in the forward and reversed processes described by 
Eq.~(\ref{eq:yt-cond}) and Eq.~(\ref{eq:xt-br-cond}) respectively.)

\end{document}